\documentclass[journal]{IEEEtran}

\usepackage{graphics} 
\usepackage{epsfig} 
\usepackage{mathptmx} 
\usepackage{times} 
\usepackage{amsmath} 
\usepackage{amssymb}  
\usepackage{booktabs}
\usepackage{multirow}
\usepackage{bbding}
\usepackage{makecell}
\usepackage{color}
\usepackage{amsfonts}
\usepackage{mathtools}
\usepackage{subcaption}
\DeclarePairedDelimiter\abs{\lvert}{\rvert}
\DeclareMathAlphabet{\mathcal}{OMS}{cmsy}{m}{n}
\usepackage{algorithm}
\usepackage{algpseudocode}

\newcommand{\etal}{\textit{et al.}}

\begin{document}
%
\title{Dual-Awareness Attention for \\Few-Shot Object Detection}
%
%
%


\author{Tung-I Chen$^1$, Yueh-Cheng Liu$^1$, Hung-Ting Su$^1$, Yu-Cheng Chang$^1$, Yu-Hsiang Lin$^1$, Jia-Fong Yeh$^1$, \\
Wen-Chin Chen$^1$, Winston H. Hsu$^{1, 2}$ 
\\ $^1$ National Taiwan University, $^2$ Mobile Drive Technology
}


%
%

\markboth{IEEE TRANSACTIONS ON MULTIMEDIA, VOL, 23, 2021}%
{Chen \etal: Dual-Awareness Attention for Few-Shot Object Detection}
%



\maketitle

\begin{abstract}
While recent progress has significantly boosted few-shot classification (FSC) performance, few-shot object detection (FSOD) remains challenging for modern learning systems.
Existing FSOD systems follow FSC approaches, ignoring critical issues such as spatial variability and uncertain representations, and consequently result in low performance.
Observing this, we propose a novel \textbf{Dual-Awareness Attention (DAnA)} mechanism that enables networks to adaptively interpret the given support images.
DAnA transforms support images into \textbf{query-position-aware} (QPA) features, guiding detection networks precisely by assigning customized support information to each local region of the query.
In addition, the proposed DAnA component is flexible and adaptable to multiple existing object detection frameworks.
By adopting DAnA, conventional object detection networks, Faster R-CNN and RetinaNet, which are not designed explicitly for few-shot learning, reach state-of-the-art performance in FSOD tasks.
In comparison with previous methods, our model significantly increases the performance by 47\% (+6.9 AP), showing remarkable ability under various evaluation settings.
\end{abstract}


%

\section{Introduction}
%
%
Few-shot object detection (FSOD) is a recently emerging and rapidly growing research topic, which has great potential in many real-world applications.
Unlike conventional object detectors, few-shot object detectors can be adapted to novel domains with only few annotated data, saving the costly data re-collection and re-training processes whenever the downstream task changes.
However, though considerable effort in recent years has been devoted, existing FSOD methods~\cite{kang2019few, yan2019meta, fan2020few, perez2020incremental, wu2020multi} suffer extremely low performance in comparison with traditional object detectors~\cite{ren2015faster, lin2017focal}.
In addition, it seems previous methods suffer performance drop not only on the novel domain but also on the base (training) domain~\cite{wang2020frustratingly, perez2020incremental}.
Moreover, by carrying out experiments, we discovered that previous FSOD models based on Faster R-CNN~\cite{ren2015faster} are incapable of reaching the performance that Faster R-CNN can achieve when being evaluated on base classes (see Tab.~\ref{tab:ZSOD_multishot}).
Though the models have been trained on a lot of annotated data, they still have difficulty in recognizing the objects they have seen.
We therefore assume the modifications associated with few-shot learning applied in prior attempts somehow undermine the ability of detection networks, resulting in limited performance.
%
%
%
%
%
%

%
\begin{figure*}[t!]
    \centering
    \includegraphics[width=\linewidth]{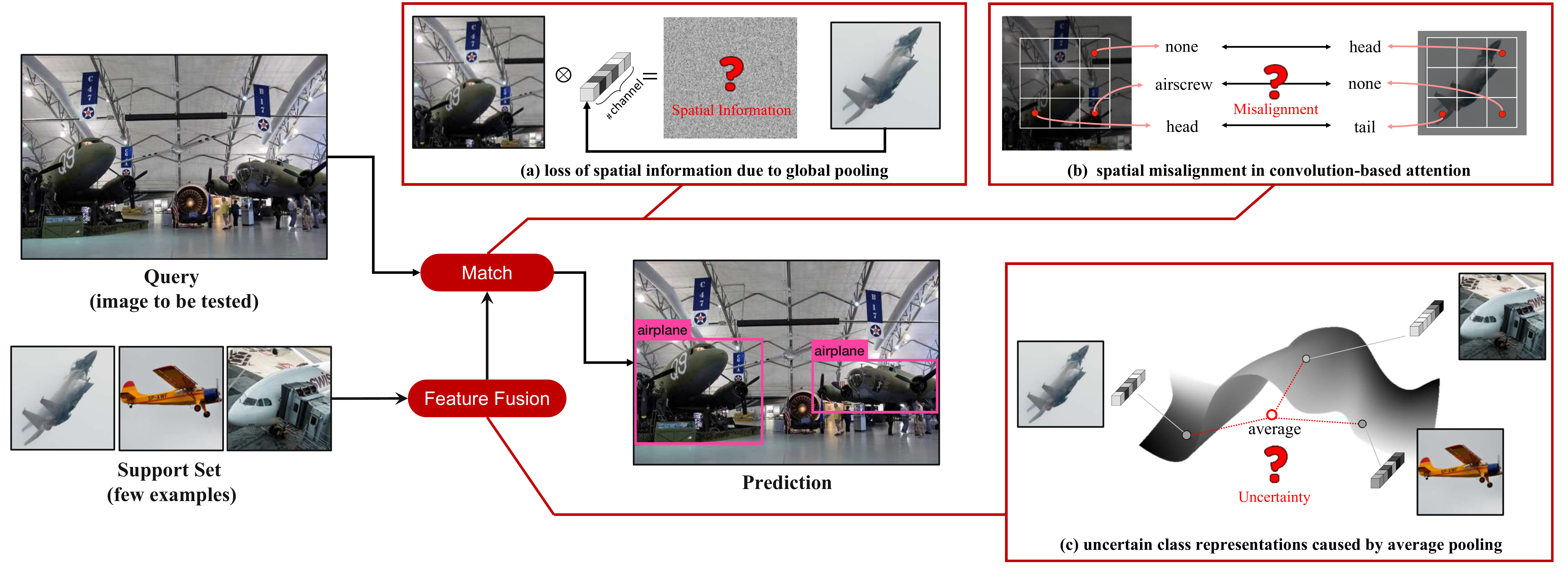}
    \caption{Illustration of few-shot object detection and the concerns in previous methods (a-c). (a) Previous works seek computational efficiency by encoding support images into global feature vectors, causing the loss of spatial and local information. (b) The convolution-based attention has difficulty in modeling relationships across objects with varying spatial distributions. (c) Performing average pooling across multiple high-level features might cause uncertainty in the resulting feature representation.}
    \label{fig:illustration}
\end{figure*}

To recast traditional object detectors into few-shot object detectors, prior works tend to leverage the methods that have been proved effective at few-shot classification (FSC).
The techniques such as building category prototypes~\cite{snell2017prototypical, karlinsky2019repmet}, ranking similarity between inputs~\cite{koch2015siamese, fan2020few} and feature map concatenation~\cite{sung2018learning, fan2020fgn} are all widely adopted.
However, unlike FSC aiming to classify images, FSOD is a much more complicated task requiring to identify and locate objects precisely.
In this work, we summarize three potential concerns that could restrict the performance of FSOD (see Fig.~\ref{fig:illustration}).
(a) First of all, prior methods~\cite{karlinsky2019repmet, kang2019few, yan2019meta, fan2020fgn, perez2020incremental, liu2020crnet} performed global pooling on support features to ensure computational efficiency.
However, the lack of spatial information would cause difficulty in measuring object-wise correlations.
(b) Secondly, convolutional neural networks (CNN) are physically inefficient at modeling varying spatial distributions~\cite{hu2019local}, so the methods using convolution-based attention~\cite{bertinetto2016fully, fan2020fgn} would suffer the same restriction.
(c) Furthermore, previous works~\cite{karlinsky2019repmet, kang2019few, yan2019meta, fan2020fgn, fan2020few, perez2020incremental} took mean features across multiple support images as class-specific representations, which heavily relies on an implicit assumption that the mean features will still be representative in the embedding space.
%
%
%

%
%
To verify the aforementioned concerns are worth studying, we conduct a pilot study with the hypothesis: \textit{If the spatial information, variability and feature uncertainty are all trivial in FSOD, the choice of support images will not severely affect the performance.}
In the experiment, the four well-trained FSOD models are tested on the same query set 100 times but the given support images will be different each time.
As shown in Fig.~\ref{fig:pilot}, even though we fix the query set, previous methods can easily be influenced by the choice of supports, resulting in a huge range of performance (up to 4.4 AP).
On the other hand, our method, which solves the concerns summarized in Fig.~\ref{fig:illustration}, achieves the highest and firmest performance.
%
%
%

%
In this work, we present a novel \textit{Dual-Awareness Attention} (DAnA) mechanism comprised of \textit{Background Attenuation} (BA) and \textit{Cross-Image Spatial Attention} (CISA) modules.
Inspired by wave interference, we propose a BA module where each feature vector ($1\times 1\times C$) of a high-level feature map ($H\times W\times C$) is viewed as a wave along the channel dimension.
By extracting the most representative feature vector from a feature map and adding it back to the feature map, those local features having different wave patterns from the extracted feature will be blurred and therefore can be easily recognized as noise by the detection network. 
Those foreground features, on the other hand, can maintain the wave patterns and be preserved after the addition.
Thus, the BA module not only attenuates irrelevant features but preserves the target information as well.
%
%

%
%
%
To determine whether two objects belong to the same class, a person might first determine the most representative features among objects ($e.g.$, dog paws, bird wings) and then make his decision according to these features.
Inspired by such a nature, we propose CISA to adaptively transform support images into various query-position-aware (QPA) vectors.
To be more specific, each QPA vector carries specific support information that is considered the most relevant to each local query region.
By measuring correlations between the query regions and their corresponding QPA vectors, the model can easily determine whether the regions should be the parts of the target object.
Also, CISA provides a more efficient and effective way to summarize information among multiple support images.
Those QPA vectors conditioned on the same query region would represent relevant information, and therefore taking the mean feature across them is more effective than previous manners.
By better utilizing support images, our method achieves the most significant improvement as the number of support images increases (see Tab.~\ref{tab:ZSOD_multishot}).
In this paper, we evaluate models across various settings, including the multi-shot, multi-way, cross-domain, and episode-based evaluations.  
In Tab.~\ref{tab:reported}, we show our method significantly outperforms the strongest baseline~\cite{xiao2020few} by 6.9 AP under the $30$-shot setting.
Furthermore, we are the first to test FSOD models on novel domains without fine-tuning (Tab.~\ref{tab:ZSOD_multishot} and Tab.~\ref{tab:ZSOD_multiway}) to further evaluate the generalization ability of each method.
We also offer a comprehensive ablation study to demonstrate the impact of each proposed component.
Our main contributions can be summarized as follows:
\begin{itemize}
    \item We point out the critical issues in previous FSOD methods that lead to the limited performance and the lack of robustness.
    \item We present novel dual-awareness attention to precisely capture object-wise correlations.
    \item We conduct comprehensive experiments to fairly assess each approach. Extensive experiments manifest the effectiveness and robustness of the proposed methodology.
\end{itemize}
%


\section{Related Works}

\subsection{Few-Shot Classification}
Few-shot classification (FSC) has multiple branches, including the optimization-based and metric-based approaches.
The optimization-based methods~\cite{ravi2016optimization, finn2017model, li2017meta, nichol2018first, lee2019meta} aim to learn a good initialization parameter set that can swiftly be adapted to new tasks within few gradient steps. 
The metric-based methods~\cite{koch2015siamese, vinyals2016matching, snell2017prototypical, sung2018learning, liu2019prototype, tian2020rethinking}, on the other hand, compute the distance between learned representations in the embedding space.
%
%
The concept of prototypical representation~\cite{snell2017prototypical} is widely adopted in FSC, which takes the mean feature over different support images as a class-specific embedding.
Such a strategy is intuitive yet the data scarcity in few-shot scenarios might lead to biased prototypes and consequently hinder the performance~\cite{liu2019prototype}.
To enhance the reliability of class representations, Tian \etal~\cite{tian2020rethinking} leveraged pre-trained embedding and showed that using good representations is more effective than applying sophisticated meta-learning algorithms.
Although the attempts have successfully boosted the performance of FSC, the progress of few-shot object detection (FSOD) is still in a very early stage.
In this work, we explore novel attention mechanisms and significantly enhance the performance of FSOD.
%
%
%
%
%
\subsection{Attention Mechanism}
The attention modules were first developed in natural language processing (NLP) to facilitate machine translation~\cite{bahdanau2014neural, luong2015effective, gehring2017convolutional}.
Recently, inspired by the great success of Transformer~\cite{vaswani2017attention}, researchers start to explore the self-attention mechanism on computer vision (CV) problems~\cite{wang2018non, zhao2018psanet, fu2019dual, cao2019gcnet, hu2019local, zhu2019empirical}, attempting to break the physical restrictions of CNN.
Wang \etal~\cite{wang2018non} presented a pioneering approach, Non-local (NL) Neural Networks, leveraging self-attention to capture long-range dependencies in an image.
Hu \etal~\cite{hu2018squeeze} adopted a self-attention function on channels, re-weighting features along the channel dimension.
Following NL, \cite{zhao2018psanet} described the features as information flows that can be bidirectionally propagated; \cite{cao2019gcnet} showed that simplifying NL block does not deteriorate its ability but rather enhance the performance; recently, Yin \etal~\cite{yin2020disentangled} succeeded in capturing better visual clues by proposing a disentangled NL block.
Emami \etal~\cite{emami2020spa} combined spatial attention with GAN to handle image-to-image translation tasks.
Li \etal~\cite{li2020spatio} explored attention in both spatial and temporal dimensions, improving the performance of video action detection.
%
%

%
The main differentiating factor between the proposed DAnA and aforementioned attention mechanisms is that DAnA can capture cross-image dependencies and interpret support images adaptively according to the given query.
It would be plausible that the idea of DAnA can be applied to other research topics aiming to capture shared attributes among images with diverse backgrounds, viewpoints and illumination conditions~\cite{zhang2015self, han2017unified, fan2021group}, yet in this work we will emphasize the application in FSOD only.
%

%
\begin{figure}[t]
    \centering
    \includegraphics[width=\linewidth]{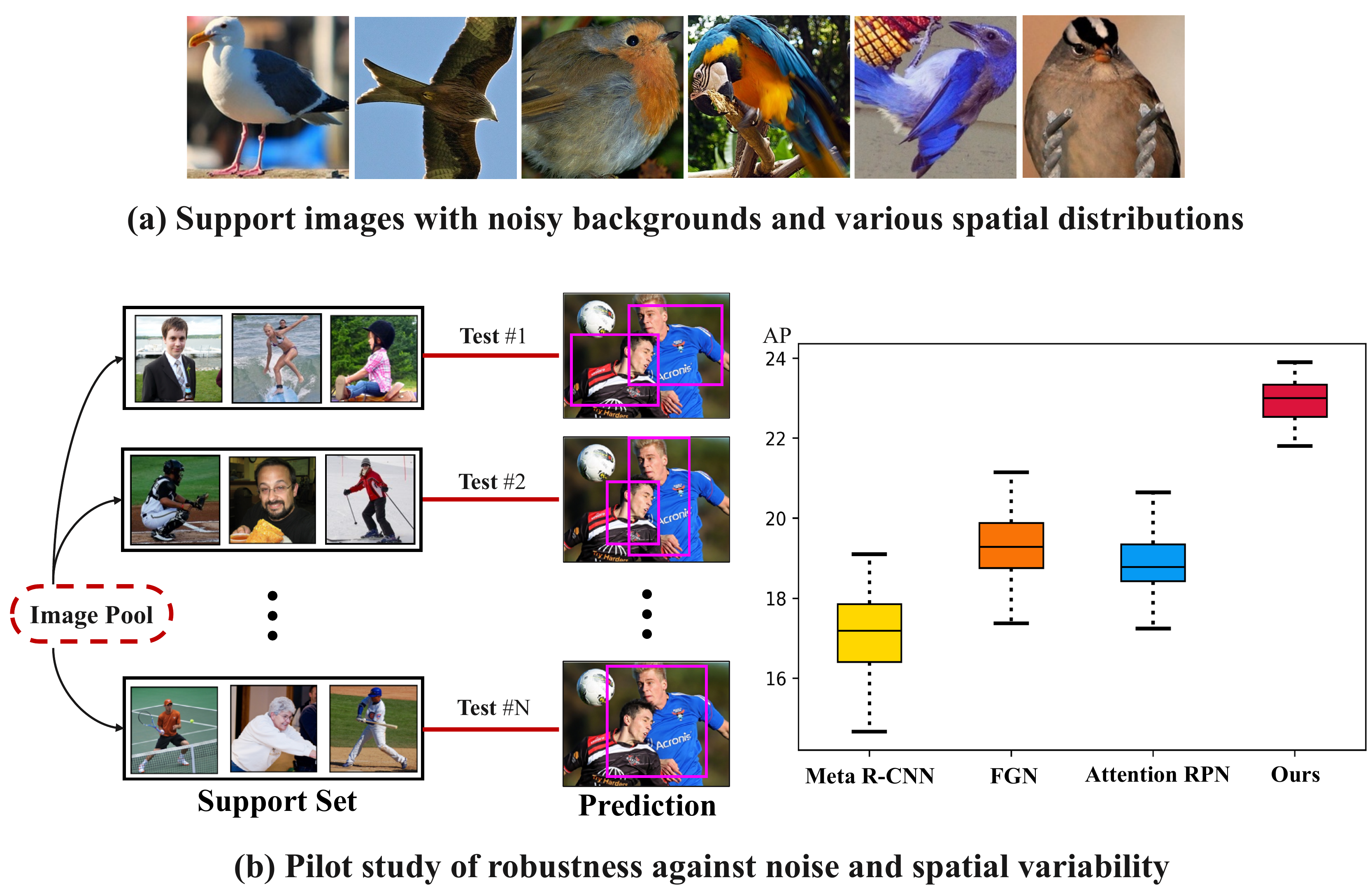}
    \caption{
    Illustration of the pilot study.
    We carry out an experiment to test whether the noise and spatial variability of support images would seriously influence FSOD results.
    The standard deviations of AP of the four methods are $[0.72, 0.81, 0.68, 0.46]$ respectively.
    Our method is effective at capturing object-wise correlations irrespective of the choice of support images.
    }
    \label{fig:pilot}
\end{figure}

\subsection{Few-Shot Object Detection}
%
Deep-learning-based object detectors have shown remarkable performance in many applications.
Two-stage detectors~\cite{girshick2014rich, girshick2015fast, ren2015faster, lin2017feature} are usually dominant in performance, and one-stage detectors~\cite{lin2017focal, law2018cornernet, tian2019fcos, zhou2019objects} are superior in run-time efficiency.
Most modern object detectors are category-specific, which means they are incapable of recognizing objects of unseen categories.
To explore generalized object detectors, previous attempts exploited transfer learning~\cite{chen2018lstd} and distance metric learning~\cite{karlinsky2019repmet} to achieve quick adaption to novel domains.
%
%
As recasting object detection problem into the few-shot learning paradigm, the idea of current methods could be somewhat similar to multiple-instance learning (MIL)~\cite{qi2007concurrent, dollar2008multiple, babenko2010robust}.
For FSOD, we can also perceive a query image as a bag, aiming to identify the positive image patches in it by capturing the contexts relevant to given support images~\cite{qi2007concurrent}.
Recently, there is an important line of works encoding support images into global vectors, measuring the similarity between feature vectors and the RoI proposals inferred by detection networks~\cite{kang2019few, yan2019meta}.
Following the spirit, \cite{fan2020fgn} perceived the task as a guided process where support features are used to guide the object detection networks.
Inspired by \cite{bertinetto2016fully}, Fan \etal~\cite{fan2020few} measured the correlations by regarding support images as kernels and performing a convolution-based operation over queries.
Current works have a tendency to take mean features as class representations and measure cross-image correlations by either feature concatenation or element-wise product~\cite{yan2019meta, karlinsky2019repmet, fan2020fgn, fan2020few}.
We argue that these techniques will engender serious issues (illustrated in Fig.~\ref{fig:illustration}) in FSOD, degenerating the performance even on the seen (training) categories.
%


\section{Methodology}
\subsection{Problem Definition}
Let $s$ be a support image and $\mathcal{S} = \{s_{i}\}^M_{i=1}$ be a support set that represents a specific category.
An individual FSOD task can be formulated as $\mathcal{T}=\{(\mathcal{S}^1, ..., \mathcal{S}^N), \mathcal{I}\}$, where $\mathcal{I}$ is a query image comprised of multiple instances and backgrounds.  
Given $\mathcal{T}$, the model should detect all the objects in $\mathcal{I}$ belonging to the target categories $\{\mathcal{S}^1, ..., \mathcal{S}^N\}$.
The object categories in a training dataset are divided into two disjoint parts: base classes $\mathcal{C}^{base}$ and novel classes $\mathcal{C}^{novel}$. 
To train a FSOD model, a meta-training set $\mathcal{H}^{train}=\{\mathcal{T}_i\}_{i=1}^{\abs{\mathcal{H}}}$ should be constructed, where all the bounding box annotations belong to $\mathcal{C}^{base}$.
Similarly, a meta-testing set $\mathcal{H}^{test}$ is constructed where all the target objects belong to $\mathcal{C}^{novel}$. 
It is allowed to use a fine-tuning set $\mathcal{H}^{finetune}$ to fine-tune the mdoel before evaluating on $\mathcal{H}^{test}$.
However, in an $N$-way $K$-shot setting, only $K$ box annotations of each novel category can be used to fine-tune the model~\cite{karlinsky2019repmet, kang2019few}.
To summarize, the primary goal of FSOD is to leverage rich source-domain knowledge in $\mathcal{H}^{train}$ to learn a model that can swiftly generalize to target domains where only few annotated data are available.
Instead of only considering the features of $\mathcal{I}$, the model $f(\mathcal{I}|\mathcal{S})$ is trained to recognize objects conditioned on the given support information.
\begin{figure}[t]
    \centering
    \includegraphics[width=\linewidth]{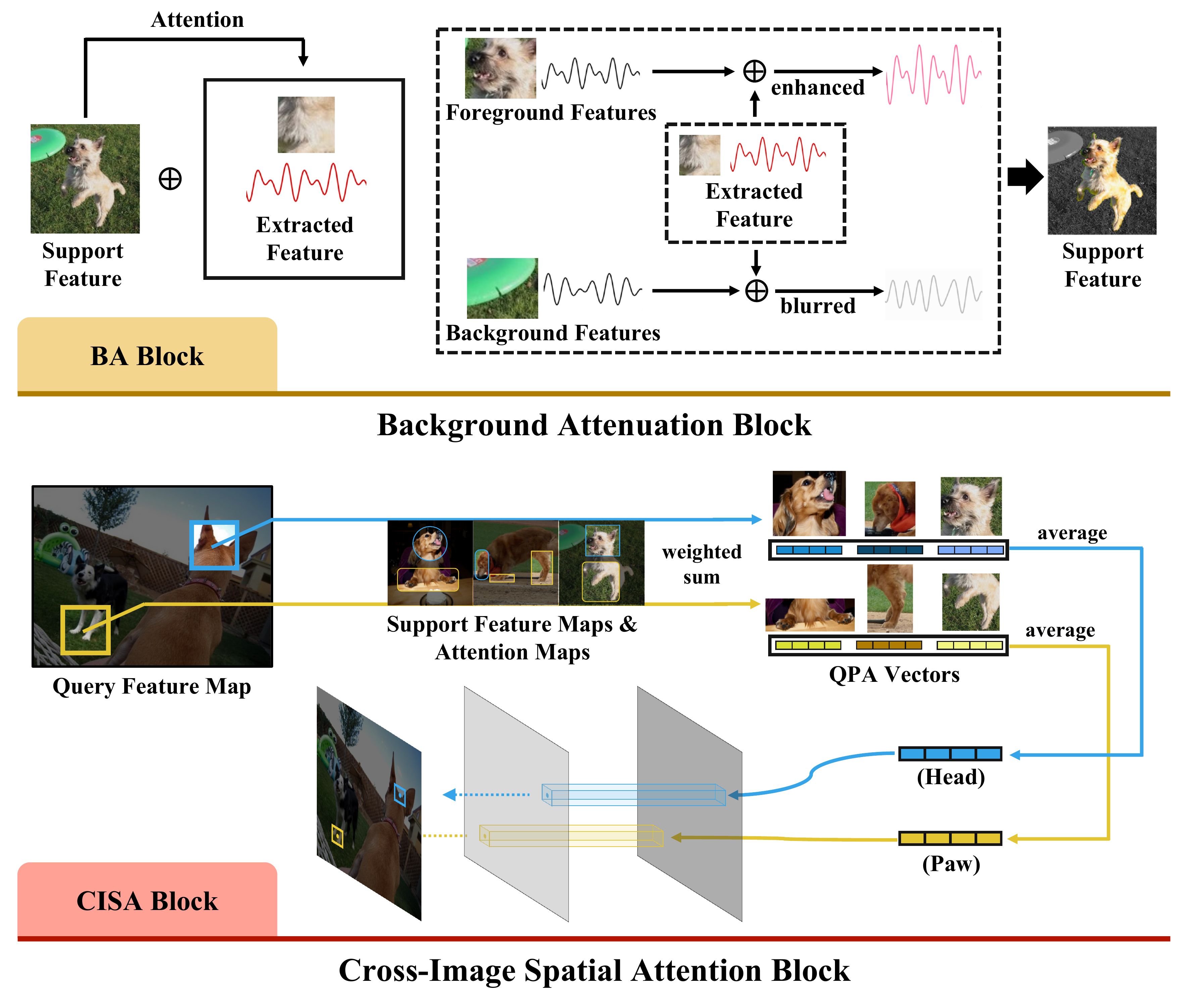}
    \caption{Illustration of our proposed methods. 
    In the BA block, the extracted feature is superimposed on the feature map to undermine irrelevant background features while maintain distinguishing information. 
    In the CISA block, support feature maps are adaptively transformed into query-position-aware (QPA) vectors, which are customized support information for each query region.
    }
    \label{fig:block_illustration}
\end{figure}
\begin{figure}[t]
    \centering
    \includegraphics[width=\linewidth]{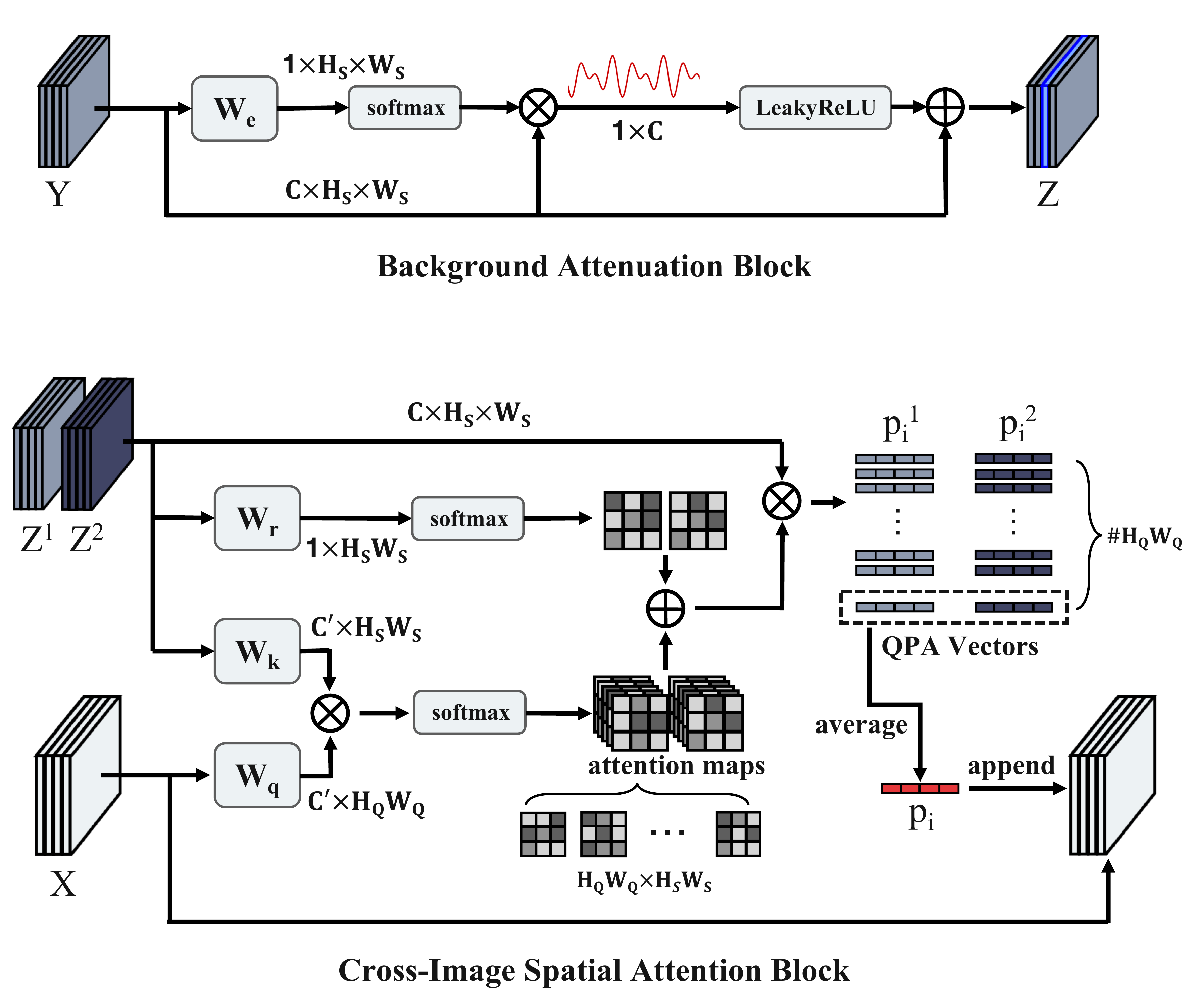}
    \caption{Detailed structures of the two proposed modules. $\otimes$ denotes matrix multiplication, and $\oplus$ denotes broadcast element-wise addition; $W$ denotes a learned weight matrix;  $Y$ and $X$ are support and query feature maps respectively; $Z^1$ denotes the $1^{st}$ feature map in the support set processed after the BA block; $p_i$ denotes the query-position-aware support vector based on the $i^{th}$ pixel of $X$.
    }
    \label{fig:blocks}
\end{figure}

\subsection{Dual-Awareness Attention}
\subsubsection{Overview}
FSOD relies on limited support information to detect novel objects, and therefore we consider two important aspects: (1) The quality of support features, and (2) how to better construct correlations between support and query images.
%
%
In this work, we propose an attention mechanism comprised of two novel modules to undermine the influence of noise and precisely measure object-wise correlations.
\subsubsection{Background Attenuation Block}
It is infeasible to always ensure high-quality support images in real-world scenarios. 
Those noise in support images will inhibit models from reaching robustness (see Fig~\ref{fig:pilot}).
We propose a novel mechanism, \textit{Background Attenuation} (BA), to undermine the irrelevant support information.
The detailed structure of BA block is illustrated in Fig.~\ref{fig:blocks}, where a support image $s$ and a query image $\mathcal{I}$ are encoded into a support feature map $Y\in\mathbb{R}^{C\times H_S\times W_S}$ and a query feature map $X\in\mathbb{R}^{C\times H_Q \times W_Q}$ by a shared CNN backbone.
In BA block, the feature map $Y$ will be reshaped and transformed by a linear learnable matrix $W_e\in\mathbb{R}^{C\times 1}$.
The process can be formulated as
\begin{equation} \label{eq:attention_in_BA}
    \mathcal{A}_\text{BA} (y_i) = \sigma (W_e y_i) = \frac{\exp(W_e y_i)}{\sum_{j\in\Omega} \exp(W_e y_j)}
\end{equation}
where $y_i\in \mathbb{R}^{1\times C}$ denotes the feature vector at $i^{\text{th}}$ pixel of $Y$; $\Omega$ is the set of all pixel indices and $\sigma$ is the softmax function applied along the spatial dimension.
Thus, the learned aggregation of $Y$ can be obtained by 
%
\begin{equation}
    G = \sum_{i\in \Omega} \mathcal{A}_\text{BA} (y_i)\cdot y_i .
\end{equation}
%
Intuitively, $G\in\mathbb{R}^{1\times C}$ should represent the most important feature of $Y$ according to the resulting attention maps.
However, we empirically discover that using the learned attention maps to filter out noise would harm the performance (see Tab.~\ref{tab:ablation}).
By visualization, we observe that only few narrow regions of support image contribute to the aggregated feature $G$.
Such a naive attention process leads to a considerable loss of support information and therefore deteriorate the ability of models.
%
%

%
Consequently, we propose a much softer attention strategy to remove noise.
In physics, \textit{interference} is a term indicating two signals superpose to form a resultant signal of greater or lower amplitudes.
Inspired by that, we view each feature vector of $Y$ as a $C$-dimensional signal and superimpose the extracted feature $G$ on those signals
\begin{equation} \label{eq:enhancement_in_SE}
    Z = Y + \alpha \cdot LeakyReLU (G)
\end{equation}
where $\alpha$ is a constant hyper-parameter, and the nonlinear function is used to rectify the results.
As illustrated in Fig.~\ref{fig:block_illustration}, if the original signals ($e.g.$, regions belonging to the dog) are aligned with $G$ along the channel dimension, the features can be enhanced or maintained.
On the contrary, those signals associated with unrelated regions ($e.g.$, the Frisbee) will be blurred after addition due to the significant difference with $G$.
Thus, the BA block can benefit detection networks by providing more discriminative support features.
It is worth mentioning the idea of BA is different from the channel-wise attention~\cite{hu2018squeeze, fu2019dual} which re-weights feature maps along the channel dimension.
In addition, unlike \cite{cao2019gcnet} constructing heavy linear transformation matrices to rew-eight semantic dependencies, BA block is more efficient since $W_e\in\mathbb{R}^{C\times 1}$ is the only learnable weights in it. 
\begin{figure}[t]
    \centering
    \includegraphics[width=\linewidth]{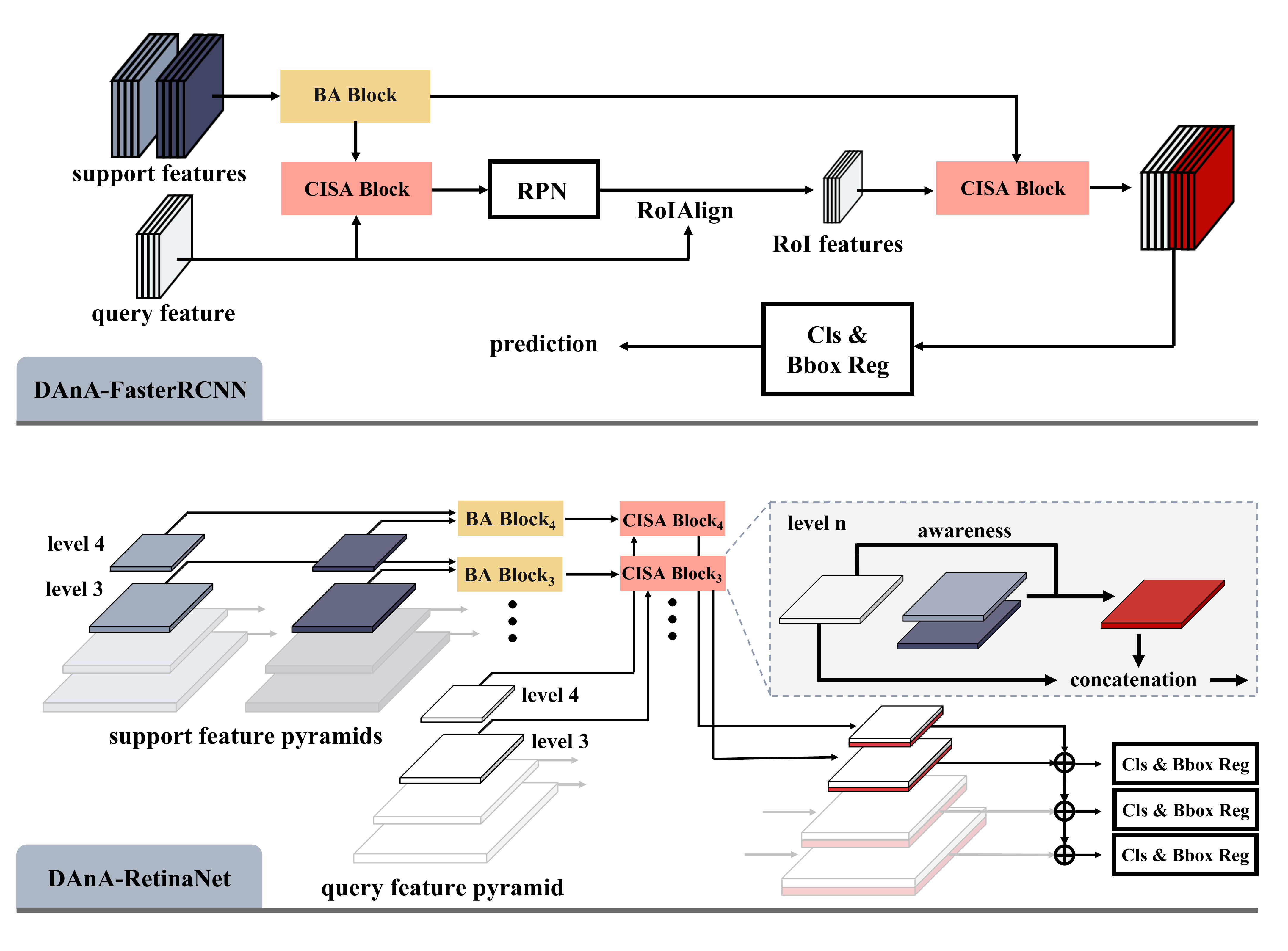}
    \caption{The model architectures of DAnA-FasterRCNN and DAnA-RetinaNet. Our proposed components are flexible and can be easily combined with existing object detection frameworks which are not originally designed for FSOD.
    }
    \label{fig:architecture}
\end{figure}

\subsubsection{Cross-Image Spatial Attention Block}
\label{section:method_cisa}
Since even the intra-class objects would have obvious deviation in appearance, the model should learn to focus on those most representative parts of the objects to determine the similarity among them.
The core idea of Cross-Image Spatial Attention (CISA) is to adaptively transform each support feature map into query-position-aware (QPA) support vectors that represent specific information of a support image.
%
%
The first step of CISA is similar to QKV attention~\cite{vaswani2017attention}, transforming $X$ and $Z$ into the query and key embeddings $\mathcal{Q}=W_{q}X, \mathcal{K} = W_{k}Z$ by learned weight matrices $W_q, W_k\in\mathbb{R}^{C\times C'}$ respectively.
The similarity scores between the query and support can then be measured by
\begin{equation} \label{eq:delta_in_dana}
    \delta(X, Z) = \sigma( (\mathcal{Q} - \mu_{\mathcal{Q}})^\top (\mathcal{K} - \mu_{\mathcal{K}}))
\end{equation}
where $\mu_{\mathcal{Q}}$ and $\mu_{\mathcal{K}}$ are the averaged embedding values over all pixels; the softmax function $\sigma$ is performed over the spatial dimension.
Furthermore, we add a simplified self-attention term~\cite{cao2019gcnet} in CISA because we assume the attention should be based on not only query-support correlations but also the support image itself.
Thus, the CISA attention function is formulated as
\begin{equation} \label{eq:attention_in_dana}
    \mathcal{A}_\text{CISA}(X, Z) = \delta (X, Z) + \beta \cdot W_r Z
\end{equation}
where $\beta$ is a constant coefficient; $W_r\in\mathbb{R}^{C\times 1}$.
%
Note that the output of $\mathcal{A}_\text{CISA}(X, Z)$ has the shape of $H_Q W_Q\times H_S W_S$, which indicates the fact that we have obtained multiple support attention maps ($H_S W_S$) conditioned on each spatial location of the query feature map ($H_Q W_Q$).
The QPA vectors can therefore be obtained by
\begin{equation} \label{eq:qpa_in_dana}
p_i = \sum_{j\in \Omega}  \mathcal{A}_\text{CISA}(X, Z)_{ij} \cdot z_j
\end{equation}
where $\Omega$ denotes all pixel indices of the support feature map.
The equation shows that all the vectors of $Z$ are adaptively weighted and aggregated into a vector $p_i\in\mathbb{R}^{1\times C}$ according to each query position $i$ (resolve Fig.~\ref{fig:illustration}~(a)).
Additionally, if there are $K$-shot images available in a support set $\{s\}^K$, we can perform average pooling across the resulting QPA vectors
\begin{equation} \label{eq:multishot_in_dana}
    p_i = \frac{1}{K} \sum_{k=1}^K p_{i, k} 
\end{equation}
without the concern of uncertainty, because those QPA vectors $p_{i, k}$ conditioned on the same entry $i$ have been refined and will carry relevant features (resolve Fig.~\ref{fig:illustration}~(c)).
Furthermore, it breaks the physical restriction of CNN since all the query features are aligned with their customized QPA vectors (resolve Fig.~\ref{fig:illustration}~(b)). 
%
%
%

\subsection{Model Architecture}
\label{section:model_architecture}
To apply the DAnA component to existing object detection frameworks, we can simply combine the output $P\in\mathbb{R}^{C\times H_Q\times W_Q}$ with $X$ into $PX\in\mathbb{R}^{2C\times H_Q\times W_Q}$ and send it to the modules such as the region proposal network (RPN) to propose regions having high responses with the given supports (see Fig.~\ref{fig:architecture}).
In our experiments, we choose Faster R-CNN~\cite{ren2015faster} and RetinaNet~\cite{lin2017focal} as the backbones to verify the effectiveness of the DAnA mechanism.
For DAnA-FasterRCNN, one CISA block is employed before RPN and the other is applied in the second stage taking cropped RoI features as inputs.
For DAnA-RetinaNet, we apply both BA and CISA modules to each level of the feature pyramids.
%

%
In addition, we follow the modification applied in \cite{fan2020few}, replacing the multi-class classification output with the binary one.
Since the pipeline of FSOD can be regarded as a matching process based on given supports, we suggest the binary output could better fit the problem scenarios where an instance will be either positively or negatively labeled.
The rest details of the proposed models remain the same as the original Faster R-CNN and RetinaNet.
%
%
%
\begin{table}[t!]
    \centering
    \resizebox{\linewidth}{!}{
        \begin{tabular}{c|c c|c c}
            \toprule
                \multirow{2}{*}{Method} & \multicolumn{2}{c|}{$10shot$} & \multicolumn{2}{c}{$30shot$} \\
                \cline{2-5}
                & $AP$ & $AP_{75}$ & $AP$ & $AP_{75}$ \\
                \hline
                TFA w/cos~\cite{wang2020frustratingly} & 10.0 & 9.3 & 13.7 & 13.4 \\
                Feature Reweighting~\cite{kang2019few} & 5.6 & 4.6 & 9.1 & 7.6 \\
                Meta R-CNN~\cite{yan2019meta} & 8.7 & 6.6 & 12.4 & 10.8 \\
                Attention RPN~\cite{fan2020few} & 11.1 & 10.6 & - & - \\
                IFSOD~\cite{perez2020incremental} & 5.1 & - & - & - \\
                MPSR~\cite{wu2020multi} & 9.8 & 9.7 & 14.1 & 14.2 \\
                Viewpoint Estimation~\cite{xiao2020few} & 12.5 & 9.8 & 14.7 & 12.2 \\
                DAnA-FasterRCNN (Ours) & \textbf{18.6} & \textbf{17.2} & \textbf{21.6} & \textbf{20.3} \\
            \bottomrule
        \end{tabular}
    }
    \caption{The performance on novel categories of COCO. All the models are trained on base categories and then fine-tuned on a small set of novel samples ($e.g.$, 10 shots of each category). After fine-tuning, models will be evaluated on the novel classes. ``-'': no reported results. 
    }
    \label{tab:reported}
\end{table}

\section{Experiments}

%
\begin{table*}[t!]
    \centering
    \resizebox{\textwidth}{!}{
        \begin{tabular}{c|ccc|ccc|ccc|ccc|ccc|ccc|c|c}
            \toprule
                \multirow{2}{*}{Method} & \multicolumn{9}{c|}{Novel Categories} & \multicolumn{9}{c|}{Base Categories} & \multirow{2}{*}{\# parameters} & \multirow{2}{*}{FPS}\\
                \cline{2-19}
                & \multicolumn{3}{c|}{$AP$} & \multicolumn{3}{c|}{$AP_{50}$} & \multicolumn{3}{c|}{$AP_{75}$} & 
                                \multicolumn{3}{c|}{$AP$} & \multicolumn{3}{c|}{$AP_{50}$} & \multicolumn{3}{c|}{$AP_{75}$} & &\\
                \hline
                Faster R-CNN$^\dagger$~\cite{ren2015faster} &
                \multicolumn{3}{c|}{ N/A } & \multicolumn{3}{c|}{ N/A } & \multicolumn{3}{c|}{ N/A } & \multicolumn{3}{c|}{ 34.3 } & \multicolumn{3}{c|}{ 58.3 } & \multicolumn{3}{c|}{ 35.6 } & $4.76 \times 10^{7}$ & 31\\
                \cline{1-19}
                \# Way & \multicolumn{3}{c|}{$1way$} & \multicolumn{3}{c|}{$1way$} & \multicolumn{3}{c|}{$1way$} & \multicolumn{3}{c|}{$1way$} & \multicolumn{3}{c|}{$1way$} & \multicolumn{3}{c|}{$1way$} &&\\
                \# Given Supports & {$1shot$} & {$3shot$} & {$5shot$} & {$1shot$} & {$3shot$} & {$5shot$} & {$1shot$} & {$3shot$} & {$5shot$} &
                {$1shot$} & {$3shot$} & {$5shot$} & {$1shot$} & {$3shot$} & {$5shot$} & {$1shot$} & {$3shot$} & {$5shot$} & & \\

                Meta R-CNN$^\dagger$~\cite{yan2019meta} & 
                8.7 & 11.1 & 11.2 &  19.9 & 25.3 & 25.9 &  6.8 & 8.5 & 8.6 &  
                27.3 & 28.6 & 28.5 &  \textbf{50.4} & \textbf{52.5} & 52.3 &  27.3 & 28.4 & 28.2    & $4.76 \times 10^{7}$ & 28\\
                FGN$^\dagger$~\cite{fan2020fgn} & 
                8.0 & 10.5 & 10.9 &  17.3 & 22.5 & 24.0 &  6.9 & 8.8 & 9.0 &  
                24.7 & 25.5 & 26.9 &  44.3 & 46.4 & 47.6 &  25.0 & 25.5 & 27.4    & $1.48 \times 10^{8}$ & 23\\
                Attention RPN$^\dagger$~\cite{fan2020few} & 
                8.7 & 10.1 & 10.6 &  19.8 & 23.0 & 24.4 &  7.0 & 8.2 & 8.3 &  
                20.6 & 22.4 & 23.0 &  37.2 & 40.8 & 42.0 &  20.5 & 22.2 & 22.4    & $1.03 \times 10^{8}$ & 21\\
                \textbf{DAnA-FasterRCNN} & \textbf{11.9} & \textbf{14.0} & \textbf{14.4} & \textbf{25.6} & \textbf{28.9} & \textbf{30.4} & \textbf{10.4} & \textbf{12.3} & \textbf{13.0} & \textbf{27.8} & \textbf{29.4} & \textbf{32.0} & 46.3 & 50.6 & \textbf{54.1} & \textbf{27.7} & \textbf{30.3} & \textbf{32.9} 
                & $1.42 \times 10^{8}$ & 24\\
                \bottomrule
        \end{tabular}
    }
    \caption{The 1-way, zero-shot evaluation on COCO. All the few-shot models are trained on base categories and then tested on novel domains without fine-tuning. Note that the performance of Faster R-CNN~\cite{ren2015faster} on base classes can serve as the upper bound because all the methods in this table leverage \cite{ren2015faster} as their backbone detector. In comparison with baselines, the relative mAP improvement of our method is up to $49\%$ on novel categories, and the performance gap with the traditional object detector has been reduced. The model size and inference speed are reported as well. $^\dagger$: re-implemented results.
    }
    \label{tab:ZSOD_multishot}
\end{table*}
\begin{table*}[t!]
    \centering
    \resizebox{\textwidth}{!}{
        \begin{tabular}{c|ccc|ccc|ccc|ccc|ccc|ccc}
            \toprule
                \multirow{2}{*}{Method} & \multicolumn{9}{c|}{Novel Categories} & \multicolumn{9}{c}{Base Categories} \\
                \cline{2-19}
                & \multicolumn{3}{c|}{$AP$} & \multicolumn{3}{c|}{$AP_{50}$} & \multicolumn{3}{c|}{$AP_{75}$} & 
                                \multicolumn{3}{c|}{$AP$} & \multicolumn{3}{c|}{$AP_{50}$} & \multicolumn{3}{c}{$AP_{75}$} \\
                \hline
                \# Way & {$1way$} & {$3way$} & {$5way$} & {$1way$} & {$3way$} & {$5way$} & {$1way$} & {$3way$} & {$5way$} & {$1way$} & {$3way$} & {$5way$} & {$1way$} & {$3way$} & {$5way$} & {$1way$} & {$3way$} & {$5way$}\\
                \# Given Supports & \multicolumn{3}{c|}{$5shot$} & \multicolumn{3}{c|}{$5shot$} & \multicolumn{3}{c|}{$5shot$} & \multicolumn{3}{c|}{$5shot$} & \multicolumn{3}{c|}{$5shot$} & \multicolumn{3}{c}{$5shot$} \\
                Meta R-CNN$^\dagger$~\cite{yan2019meta} & 
                11.2 & 11.0 & 10.2 & 25.9 & 25.2 & 23.4 & 8.6 & 8.6 & 8.0 & 28.5 & 27.4 & 26.2 & 52.3 & 50.8 & 48.7 & 28.2 & 26.8 & 25.7 \\ 
                FGN$^\dagger$~\cite{fan2020fgn} & 
                10.9 & 10.8 & 9.6 & 24.0 & 23.4 & 21.2 & 9.0 & 9.1 & 8.1 & 26.9 & 25.1 & 23.6 & 47.6 & 45.4 & 42.4 & 27.4 & 25.3 & 23.8 \\ 
                Attention RPN$^\dagger$~\cite{fan2020few} & 
                10.6 & 9.8 & 9.0 & 24.4 & 22.8 & 20.8 & 8.3 & 7.9 & 7.3 & 23.0 & 21.3 & 20.2 & 42.0 & 39.8 & 37.7 & 22.4 & 20.6 & 19.7 \\

                \textbf{DAnA-FasterRCNN} & 
                \textbf{14.4} & \textbf{13.7} & \textbf{12.6} & \textbf{30.4} & \textbf{28.2} & \textbf{25.9} & \textbf{13.0} & \textbf{12.4} & \textbf{11.3} & \textbf{32.0} & \textbf{31.0} & \textbf{29.5} & \textbf{54.1} & \textbf{52.2} & \textbf{49.9} & \textbf{32.9} & \textbf{31.7} & \textbf{30.3} \\
                \bottomrule
        \end{tabular}
    }
    \caption{The multi-way, zero-shot evaluation on COCO. Under the $N$-way setting, we introduce a support set comprised of $N$ categories to the model and therefore the model should detect the objects belonging to any of these $N$ categories.
    }
    \label{tab:ZSOD_multiway}
\end{table*}
\begin{figure*}[t!]
    \centering
    \includegraphics[width=\linewidth]{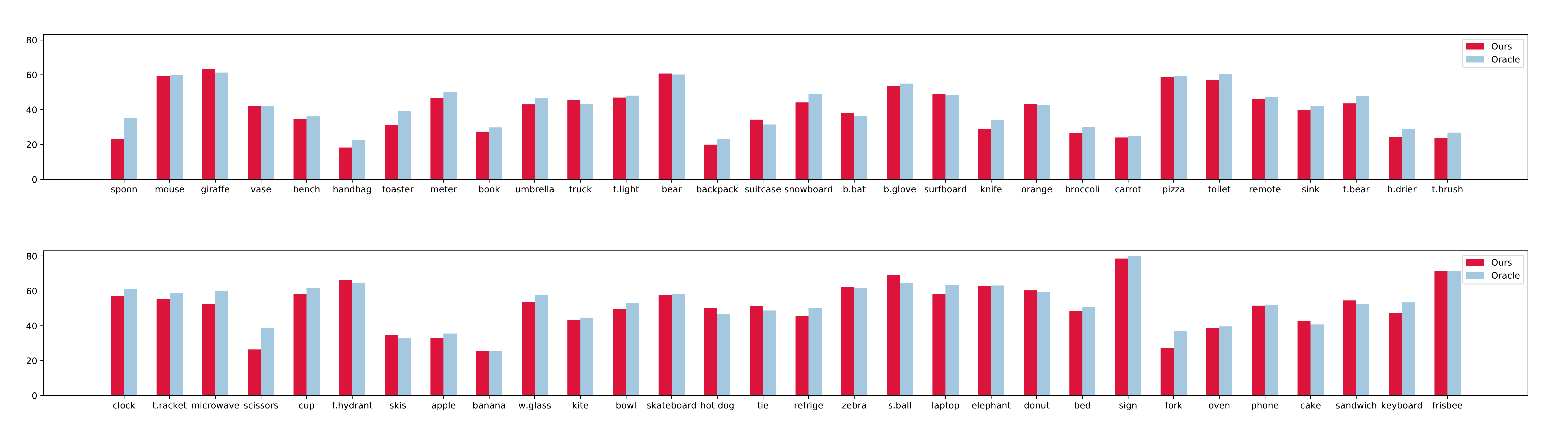}
    \caption{Visualization of the AP on each base (training) category of COCO. A well-developed few-shot object detector should be comparable to the traditional object detector on the training categories if it has not been fine-tuned. This figure demonstrates our model indeed achieves competitive performance on training categories and even surpasses the traditional one on particular classes.
    }
    \label{fig:base_performance}
\end{figure*} 

\subsection{Implementation Details}
By default, we employ the pretrained ResNet50 as the feature extractor, and the batch size is set to 32 for all the experiments.
All the models including baselines were implemented in Pytorch~\cite{paszke2017automatic} on a workstation with 4 NVIDIA Tesla V100.
The shorter side of query images is resized to 600 pixels, while the longer side is cropped at 1000. 
Each support image will be zero-padded and then resized to a square image of 320 × 320.
The embedding dimension $C'$ in the weight matrices $W_{q}$ and $W_{k}$ is a quarter of the number of original feature channels $C$. 
The constant coefficients $\alpha$ and $\beta$ of the proposed modules are 0.5 and 0.1 individually.
For those models based on Faster R-CNN~\cite{ren2015faster}, we adopt SGD with an initial learning rate of 0.001, which decays into 0.0001 after 12 epochs; the four anchor scales are [$60^2$, $120^2$, $240^2$, $480^2$] and the three aspect ratios are [0.5, 1.0, 2.0]; the momentum and weight decay coefficients are set to 0.9 and 0.0005.
For DAnA-RetinaNet, we construct a pyramid with levels $P3$ through $P7$ ($P^{l}$ has resolution $2^{l}$ lower than the input) and adopt Adam with a learning rate of 0.00001.
We adopt the two-way contrastive training strategy, which is first proposed in \cite{fan2020few}, to train the baselines and our models.

\subsection{Experimental Settings}
\subsubsection{Dataset}
\label{section:dataset}
In the experiments, we leverage the challenging Microsoft COCO 2017~\cite{lin2014microsoft} benchmark for evaluation.
COCO has 80 object classes, consisting of a training set with 118,287 images and a validation set with 4,952 images.
Generally, the validation split of COCO will serve as the testing data due to the fact that the testing split is not released to the public.
We define the 20 COCO categories intersecting with PASCAL VOC~\cite{everingham2010pascal} as the novel classes, while the rest 60 categories covered by COCO but not VOC to be the base classes.

\subsubsection{FSOD Training}
Following the paradigm of meta-learning, the training data are organized into \textit{episodes}.
Each episode contains one or more support sets, a query image and corresponding ground truth annotations.
Unlike the queries in FSC, a query image in FSOD might contain multiple objects of different categories.
Therefore, in each training step we will choose one of the object categories as the target class.
A support set of the target class will be offered and the model should learn to detect the target objects conditioned on the given support information.
Note that all the bounding box annotations belonging to the novel classes have been removed in order the prevent models from peeking at the novel task.

\subsubsection{FSOD Evaluation}
\label{section:FSOD_Evaluation}
Generally, \textit{N-way K-shot} in the few-shot paradigm indicates that we can use $K$ images of each novel category to adapt the models before testing. 
However, we observe previous works adopted different evaluation settings and fine-tuning protocols, leading to the ambiguity in evaluation.
For instance, since a query in FSOD might consist of multiple objects of different categories, Kang \etal~\cite{kang2019few} defined that there should be only $K$ box annotations of each category in the fine-tuning set, while Chen \etal~\cite{chen2018lstd} adopted the protocol of collecting $K$ query images of each category to fine-tune their models.
To thoroughly evaluate the proposed method, in Tab.~\ref{tab:reported} we follow the widely-used protocol in \cite{kang2019few} and compare our results with those works adopting the same setting.
Furthermore, we carry out additional experiments under our proposed zero-shot protocol (Tab.~\ref{tab:ZSOD_multishot},~\ref{tab:ZSOD_multiway},~\ref{tab:pascal2coco},~\ref{tab:episode}), where we do NOT leverage any annotated data of the novel classes to fine-tune models.
We expect a well-developed few-shot object detector can deal with novel objects as long as few support images are given, reducing the costly data re-collection and annotations processes. 
%
%

\subsection{Generic FSOD Protocol}
The performance of each method under the general FSOD protocol is reported in Tab.~\ref{tab:reported}.
In this table, we comply with the evaluation protocol widely used in previous works and compare our results with the numbers reported in previous papers.  
As we have discussed in Sec.~\ref{section:dataset}, the models will not be exposed to the annotations of the 20 novel classes during training.
After training, there will be $K=10, 30$ bounding box annotations per novel class can be used to fine-tune the models.
Under such a challenging setting, Tab.~\ref{tab:reported} shows our method significantly outperforms the other baselines.
By equipping with DAnA, Faster R-CNN conspicuously outperforms the strongest baseline~\cite{xiao2020few} by 6.1 and 6.9 AP under the $10$-shot and $30$-shot settings respectively.
\begin{table}[t!]
    \centering
    \resizebox{\linewidth}{!}{
        \begin{tabular}{c|ccc|ccc}
                \toprule
                \multirow{2}{*}{Method} & \multicolumn{3}{c|}{Novel Categories} & \multicolumn{3}{c}{Base Categories} \\
                \cline{2-7}
                & \multicolumn{3}{c|}{$AP$} & \multicolumn{3}{c}{$AP$} \\
                \hline
                \# Way & \multicolumn{3}{c|}{$1way$} & \multicolumn{3}{c}{$1way$} \\
                \# Given Supports & {$1shot$} & {$3shot$} & {$5shot$} & {$1shot$} & {$3shot$} & {$5shot$} \\
                Meta R-CNN$^\dagger$~\cite{yan2019meta} & 
                7.8 & 9.0 & 9.3 & 33.7 & 34.8 & 35.1 \\
                FGN$^\dagger$~\cite{fan2020fgn} & 
                6.7 & 7.4 & 7.7 & 30.5 & 31.2 & 32.1 \\
                Attention RPN$^\dagger$~\cite{fan2020few} & 
                6.8 & 7.4 & 7.7 & 23.7 & 26.6 & 27.2  \\
                \textbf{DAnA-RetinaNet} &
                9.1 & 10.3 & 11.5 & 36.5 & 36.9 & 37.2\\
                \textbf{DAnA-FasterRCNN} & 
                \textbf{10.0} & \textbf{11.6} & \textbf{11.9} & \textbf{36.5} & \textbf{38.1} & \textbf{38.6} \\
                \bottomrule
            \end{tabular}
    }
    \caption{
    The results of PASCAL2COCO evaluation. All the models are trained on PASCAL VOC 2007 and tested on COCO 2014. In this setting, the base domain denotes the 20 classes shared between PASCAL VOC and COCO, while the novel domain denotes the other 60 classes in the COCO dataset.
    }
    \label{tab:pascal2coco}
\end{table}

\subsection{Zero-shot FSOD Protocol}
In this section, we evaluate the models under a more challenging setting without fine-tuning.
Note that the word ``zero-shot'' here is not referred to the zero-shot learning paradigm~\cite{socher2013zero} but referred to the fact that we do not take any novel image to fine-tuned the models.
In order to evaluate each method under an unified protocol, we re-implement three previous methods: Meta R-CNN~\cite{yan2019meta}, FGN~\cite{fan2020fgn} and Attention RPN~\cite{fan2020few}.
These baselines and our model are all based on Faster R-CNN, and therefore we can fairly analyze the impact brought by different attention mechanisms.
Note that we have slight modifications on these methods: 1) For the model architectures, we replace the multi-class classification output with the binary one as explained in \ref{section:model_architecture}. 2) For training, we apply the two-way contrastive strategy~\cite{fan2020few} to each method since we empirically observe it can improve the performance.
In addition, as we discussed in Sec.~\ref{section:FSOD_Evaluation}, all the methods are evaluated without fine-tuning under such a protocol.
Tab.~\ref{tab:ZSOD_multishot} shows the results predicted with different numbers of support images.
With 5-shot support images given at inference, DAnA-FasterRCNN outperforms FGN and Attention RPN by $3.5/6.4/4.0$ and $3.8/6.0/4.7$ mAP respectively on $AP/AP_{50}/AP_{75}$ metrics.
Moreover, the proposed model achieves the most significant improvement as the shot increases, supporting our argument that DAnA can better retrieve the information carried within support images by leveraging QPA features.
Intuitively, without being fine-tuned on the novel domain, a well-developed FSOD model should perform well on base classes.
In order to evaluate to what extent the ability of detection networks has been undermined by the modifications, we also provide the performance of Faster R-CNN~\cite{ren2015faster}, which is the backbone of these few-shot object detectors. 
The Faster R-CNN is trained on the 60 base classes and has not been exposed to the novel categories.
The performance of it can be regarded as the upper bound of performance on base classes.
By precisely capturing object-wise correlations, our model has a much smaller gap with the upper bound comparing with baselines.
%
%
Additionally, we report the model sizes and inference speed in Tab.~\ref{tab:ZSOD_multishot} as well.
The proposed DAnA component only causes little additional cost in comparison to prior approaches.
\begin{table}[t!]
    \centering
    \resizebox{\linewidth}{!}{
        \begin{tabular}{c|ccc|ccc}
            \toprule
            \multirow{2}{*}{Method} & \multicolumn{6}{c}{Novel Categories} \\
            \cline{2-7}
            & \multicolumn{3}{c|}{$AP$} & \multicolumn{3}{c}{$AP_{75}$} \\
            \hline
            \# Way & \multicolumn{3}{c|}{$1way$} & \multicolumn{3}{c}{$1way$} \\
            \# Given Supports & {$1shot$} & {$3shot$} & {$5shot$} & {$1shot$} & {$3shot$} & {$5shot$} \\
            Meta R-CNN$^\dagger$~\cite{yan2019meta} & 14.7 & 17.0 & 17.4 & 13.1 & 15.0 & 15.2 \\
            FGN$^\dagger$~\cite{fan2020fgn} & 14.9 & 16.7 & 17.9 & 13.8 & 15.5 & 16.8 \\
            Attention RPN$^\dagger$~\cite{fan2020few} & 15.0 & 17.1 & 18.1 & 13.7 & 15.4 & 16.2 \\
            \textbf{DAnA-RetinaNet} & 16.6 & 18.8 & 19.5 & 15.5 & 17.7 & 18.4 \\
            \textbf{DAnA-FasterRCNN} & \textbf{17.9} & \textbf{21.3} & \textbf{21.6} & \textbf{17.0} & \textbf{20.2} & \textbf{20.4} \\
            \bottomrule
        \end{tabular}
    }
    \caption{
    Following \cite{karlinsky2019repmet}, we construct class-balanced episodes from COCO for evaluation. In addition, we report the performance of two different detection networks adopting the proposed DAnA component. Even though the performance of DAnA-RetinaNet is less remarkable than DAnA-FasterRCNN, it still significantly outperforms the baselines which are based on Faster R-CNN, showing the effectiveness of our proposed attention mechanism.
    }
    \label{tab:episode}
\end{table}
In Tab.~\ref{tab:ZSOD_multiway}, we further evaluate each method under the multi-way evaluation setting.
Under the multi-way setting, models must detect more than one object category conditioned on the given support set. 
Suppose there are $M$ categories in a query image, we will randomly sample $N$ classes from them to form an $N$-way support set.
Sometimes the number of object classes in a query image will be less than $N$, and we just select other $N - M$ classes to form an $N$-way support set.
Fan \etal~\cite{fan2020fgn} evaluated models under the COCO2VOC setting, where models are trained on the 60 COCO categories disjoint with VOC, and then tested on PASCAL VOC~\cite{everingham2010pascal}.
In this work, we consider a much more challenging setting termed PASCAL2COCO, where models are trained on PASCAL VOC 2007 and tested on the COCO benchmark.
The 60 categories disjoint with VOC will serve as the novel categories, which is opposite to the setting in \cite{fan2020fgn}. 
In Tab.~\ref{tab:pascal2coco}, even the models are applied to a domain that is more complicated than the training domain, the models equipped with DAnA still demonstrate remarkable ability.
In Tab.~\ref{tab:episode}, we apply the same zero-shot protocol but follow the tradition in FSC to construct episodes.
Instead of directly testing by COCO validation split, we follow the episode-based evaluation protocol defined in RepMet~\cite{karlinsky2019repmet} to prepare 500 random evaluation episodes in advance.
In each episode, take $N$-way $K$-shot evaluation for example, there will be $K$ support images and 10 query images for each of the $N$ categories.
Consequently, there will be $K\times N$ support images and $10\times N$ query images in each episode.
In Tab.~\ref{tab:pascal2coco} and Tab.~\ref{tab:episode}, we have compared DAnA-FasterRCNN with another proposed framework, DAnA-RetinaNet, to evaluate the performance as adopting different detection networks. 
Generally, the two-stage few-shot object detectors~\cite{wang2020frustratingly, yan2019meta, fan2020few, wu2020multi, xiao2020few} with region proposal network (RPN) have higher performance than the one-stage~\cite{kang2019few, perez2020incremental} methods (see Tab.~\ref{tab:reported}).
Nonetheless, though our DAnA-RetinaNet is based on the one-stage object detector, it still outperforms previous methods using two-stage networks, which suffices to verify the effectiveness and dexterity of our DAnA.
On the other hand, it seems the use of feature pyramids does not help DAnA-RetinaNet reach better results.
The result could be caused by the lack of RPN.
Without RPN, networks must tackle a nearly exhaustive list of potential object locations.
Furthermore, in FSOD, those foregrounds irrelevant to the presented support set should be classified as backgrounds as well.
Therefore, the huge amount of anchors and the extremely imbalanced foreground-background ratio would seriously undermine the ability of DAnA-RetinaNet.
To conclude, we suggest RPN plays a deep role in fully presenting the advantages of DAnA, which is verified by the aforementioned experiments and the ablation study.
%

%
\begin{table}[t!]
    \centering
    \resizebox{\linewidth}{!}{
        \begin{tabular}{ccccccccccc}
            \toprule
                & Architecture & Average Pooling & \multicolumn{2}{c}{Denoising} & \multicolumn{2}{c}{Combination} & \multicolumn{2}{c}{Novel Categories} &\\
                \cmidrule(lr){2-2}\cmidrule(lr){3-3}\cmidrule(lr){4-5}\cmidrule(lr){6-7}\cmidrule(lr){8-9}
                 & RPN & across QPA features & Mask & BA & product & concat & $AP$ & $AP_{50}$\\
                \midrule
                a              & \checkmark &    &&     & & \checkmark                   & 12.6 & 26.9\\
                b             & \checkmark & \checkmark     &&     & & \checkmark                       & 13.5 & 28.4\\
                c             & \checkmark & \checkmark     &&     & \checkmark &                       & 11.4 & 24.2\\
                d      & \checkmark & \checkmark   &\checkmark &  & & \checkmark          & 10.5 & 22.8\\
                e     &  & \checkmark   &&\checkmark    & & \checkmark             & 11.3 & 22.3\\
                f     & \checkmark & \checkmark   &&\checkmark    & & \checkmark             & \textbf{13.8} & \textbf{28.8}\\
            \bottomrule
      \end{tabular}
    }
    \caption{The results of ablations. (a,b) shows the effectiveness of using QPA features. (b,f) and (d,f) show the effectiveness of the proposed BA mechanism. (e,f) shows RPN is a critical component to fully represent the advantages of DAnA.
    }
    \label{tab:ablation}
\end{table}
\begin{figure*}[t]
    \centering
    \includegraphics[width=\linewidth]{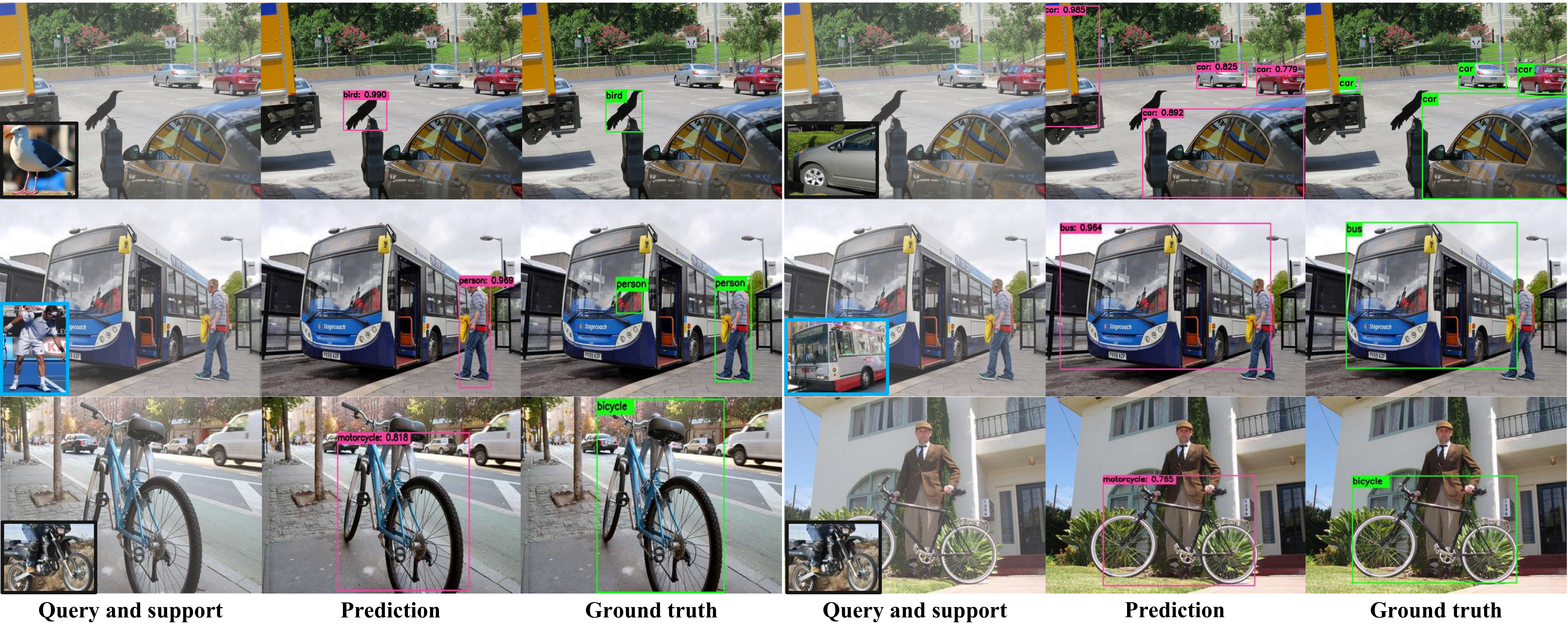}
    \caption{At training, the model has never seen the novel categories, including bird, car, dog, person, motorcycle, bicycle, etc. However, given different support examples, our model is capable of recognizing unseen target objects in the query image. The last row represents the failure case where the model confuses a bicycle with a motorcycle.
    }
    \label{fig:prediction}
\end{figure*} 
\begin{figure*}[h]
    \centering
    \includegraphics[width=\linewidth]{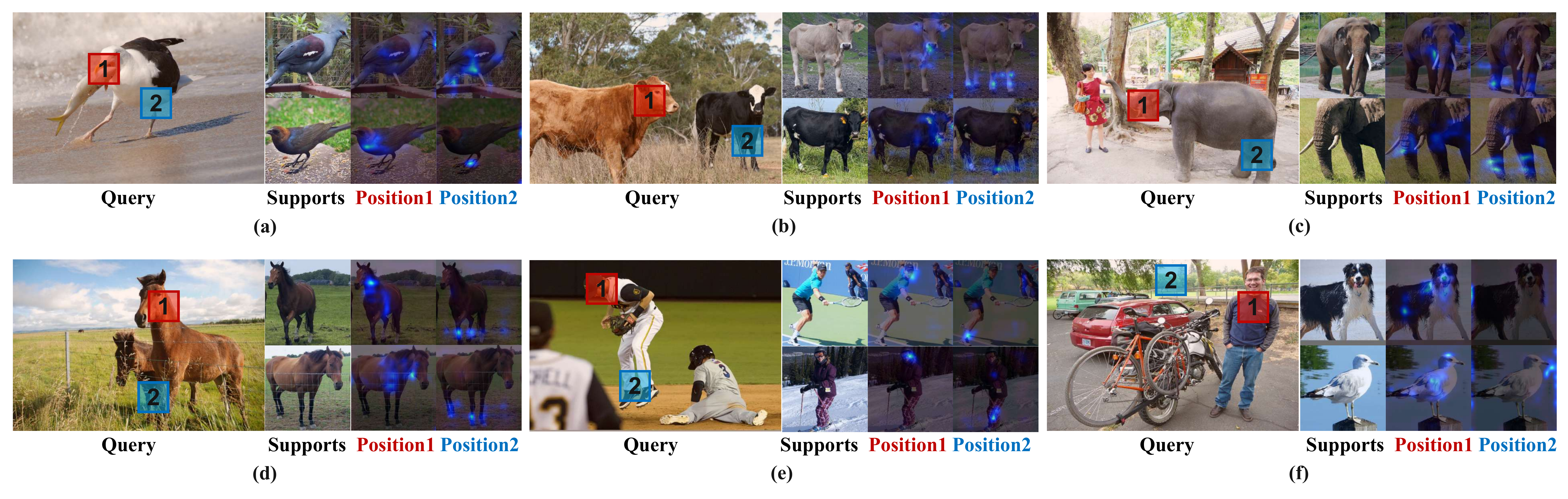}
    \caption{For each query image, we consider two different regions (colored in red and blue) and visualize their corresponding support attention maps. The CISA module is capable of capturing the semantic correspondence between the query and support images ($e.g.$, the head or legs of an animal). (f) represents the case where neither of the support categories exists in the query image.
    }
    \label{fig:attention_visualization}
\end{figure*} 

\subsection{Ablations}
In Tab.~\ref{tab:ablation}, we show the results of the ablation study to analyze the impact brought by each component.
\paragraph{Whether CISA is effective at utilizing support images}
In Sec.~\ref{section:method_cisa}, we argue that the issue of feature uncertainty can be addressed by performing average pooling across query-position-aware (QPA) vectors.
In the ablations, we investigate the improvement brought by our CISA component.
For setting (a), we directly take the mean feature over multiple images as the class representations.
The ablation (a, b) suffices to show that transforming ordinary features into QPA vectors indeed brings improvement on the performance.
\paragraph{Why not just denoise by soft attention}
To attenuate noise in support images, it is intuitive to directly learn a soft attention mask and reweights the importance of each pixel.
In Tab.~\ref{tab:ablation}, ``Mask'' means we leverage CNN to learn a soft attention mask and perform element-wise product between the mask and the support feature map.
As it can be observed, the result suggests that using the BA module is more effective than using general soft attention by preserving much more information.
\paragraph{How to combine query features and QPA vectors} 
Since the resulting QPA features will have the same size as query feature maps, they can be combined by either concatenation or element-wise product.
According to (b, c), we conclude that concatenation is a better strategy for DAnA.
\paragraph{The importance of RPN}
In the experiments, we have compared the two proposed detection networks, DAnA-FasterRCNN (with RPN) and DAnA-RetinaNet (without RPN).
Based on the results in Tab.~\ref{tab:pascal2coco} and \ref{tab:episode}, we conclude that RPN is a critical component in our problem.
Nonetheless, there could be other factors influencing the results, such as the use of feature pyramids and Focal Loss~\cite{lin2017focal}.
Therefore, in the ablations (e), we re-train a DAnA-RetinaNet but remove the feature pyramids from it.
However, we still adopt the Focal Loss in (e) because either RPN or Focal Loss is necessary to prevent the model from being severely deteriorated by numerous negative anchors.
The result (e, f) shows that, without the influence of feature pyramids, the model including RPN has significantly better performance.
To conclude, we suggest RPN is an indispensable component to fully represent the effectiveness of DAnA.
%

%

%

\subsection{Visualizations}
The examples of few-shot object detection are demonstrated in Fig.~\ref{fig:prediction}.
In FSOD, the prediction is dependent on the object category of the given support set.
Notably, all the target instances in Fig.~\ref{fig:prediction} belong to the novel classes, so the model has never been trained to recognize these objects.
However, given few support images, the proposed model is capable of recognizing and locating the instances.
The last row of Fig.~\ref{fig:prediction} presents a failure case where a motorcycle is given yet the bicycle in the query is detected.
We also visualize the cross-image spatial attention (CISA) in Fig.~\ref{fig:attention_visualization}.
The CISA module is capable of capturing the semantic correspondence between the query and support.
Take (e) for example, given the head (colored in red) or the feet of a human (colored in blue), the attention maps will highlight the head or the feet areas of the support images.
The result (f) represents the case that neither of the support categories exists in the query image, which shows that if there is no corresponding contextual information, CISA has a tendency to highlight the most distinguishing components that can best describe an object.

\section{Conclusion}
In this work, we observed the problem of spatial misalignment and feature uncertainty in the challenging few-shot object detection (FSOD) task. We propose a novel and effective Dual-Awareness Attention (DAnA) mechanism to tackle the problem. Our method is adaptable to both one-stage and two-stage object detection networks. DAnA remarkably boosts the FSOD performance on the COCO benchmark and reaches state-of-the-art results. We are excited to point out a new direction to solve FSOD tasks. We encourage future works to extend our method to other challenging tasks such as few-shot instance segmentation and co-salient object detection.
%


%



\section*{Acknowledgment}
This work was supported in part by the Ministry of Science and Technology, Taiwan, under Grant MOST 110-2634-F-002-026. We benefit from NVIDIA DGX-1 AI Supercomputer and are grateful to the National Center for High-performance Computing.
%






%




\bibliographystyle{IEEEtran}
\bibliography{main}

%






\begin{IEEEbiography}
    [{\includegraphics[width=1in,height=1.25in,clip,keepaspectratio]{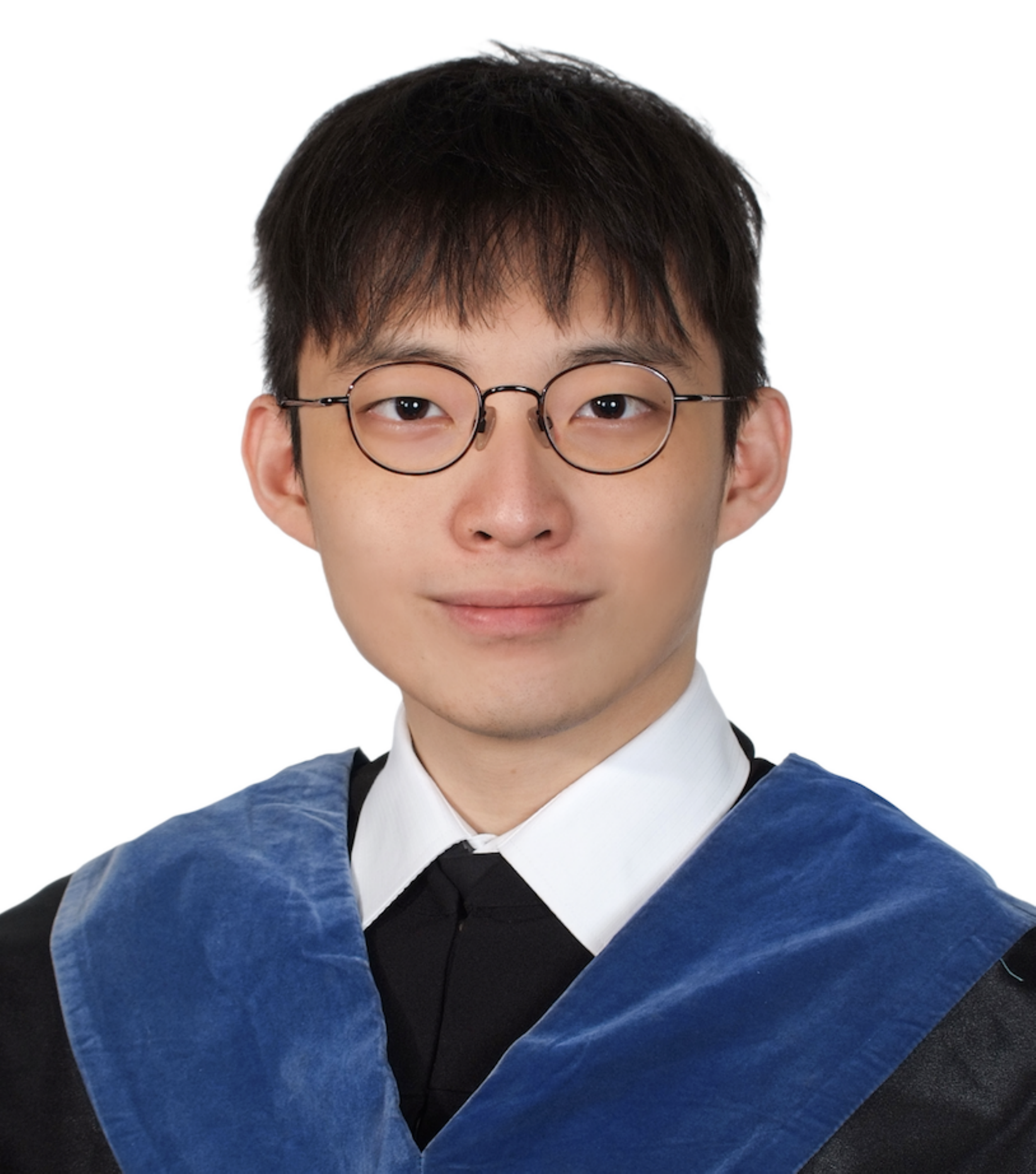}}]{Tung-I Chen}
received the B.E. degree in the Department of Biomedical Engineering, National Cheng Kung University (NCKU) in 2019, and the M.S. degree in the Department of Computer Science \& Information Engineering, National Taiwan University (NTU) in 2021. His research interests include computer vision, object detection and machine learning theory.
\end{IEEEbiography}

\begin{IEEEbiography}
    [{\includegraphics[width=1in,height=1.25in,clip,keepaspectratio]{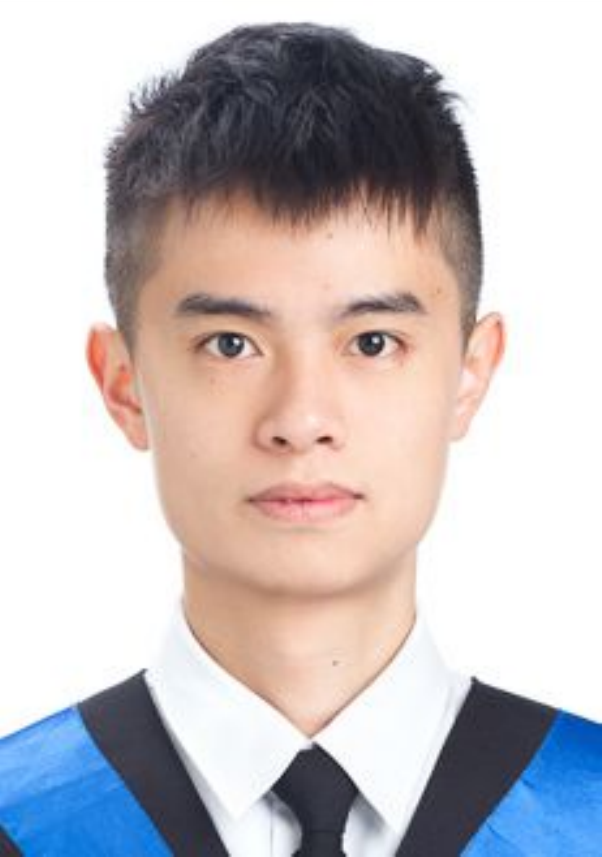}}]{Yueh-Cheng Liu}
received M.S. degree from computer science department in National Taiwan University (NTU) in 2020. He is currently a research assistant in vision science lab in EE department at National Tsing Hua University. His research interests include computer vision, robotic learning, and 3D scene understanding.
\end{IEEEbiography}

\begin{IEEEbiography}
    [{\includegraphics[width=1in,height=1.25in,clip,keepaspectratio]{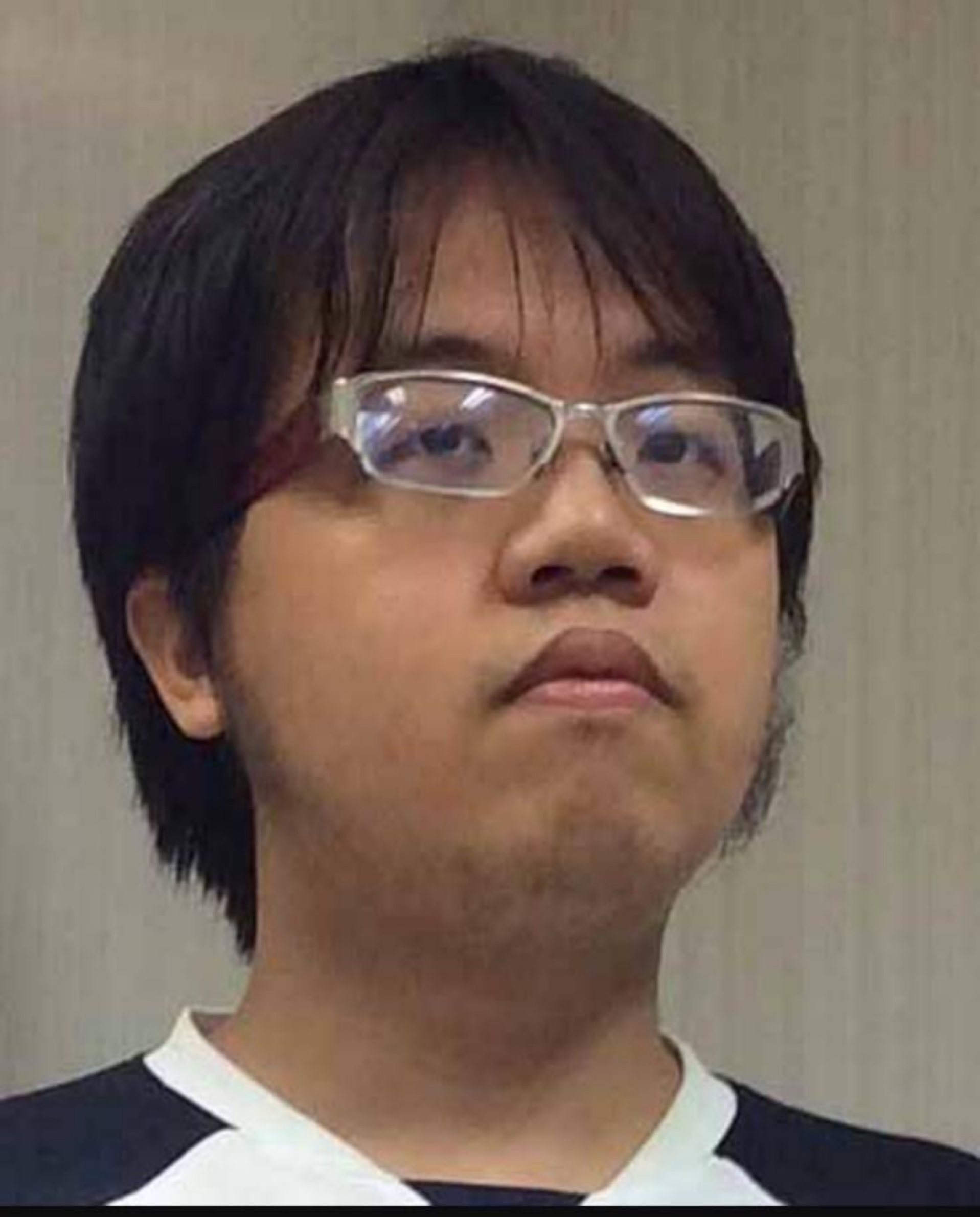}}]{Hung-Ting Su}
is currently pursuing the Ph.D. degree with the Graduated Institute of Networking and Multimedia, National Taiwan University, Taipei, Taiwan. His current research interests include multi-modal comprehension and unsupervised learning.
\end{IEEEbiography}

\begin{IEEEbiography}
    [{\includegraphics[width=1in,height=1.25in,clip,keepaspectratio]{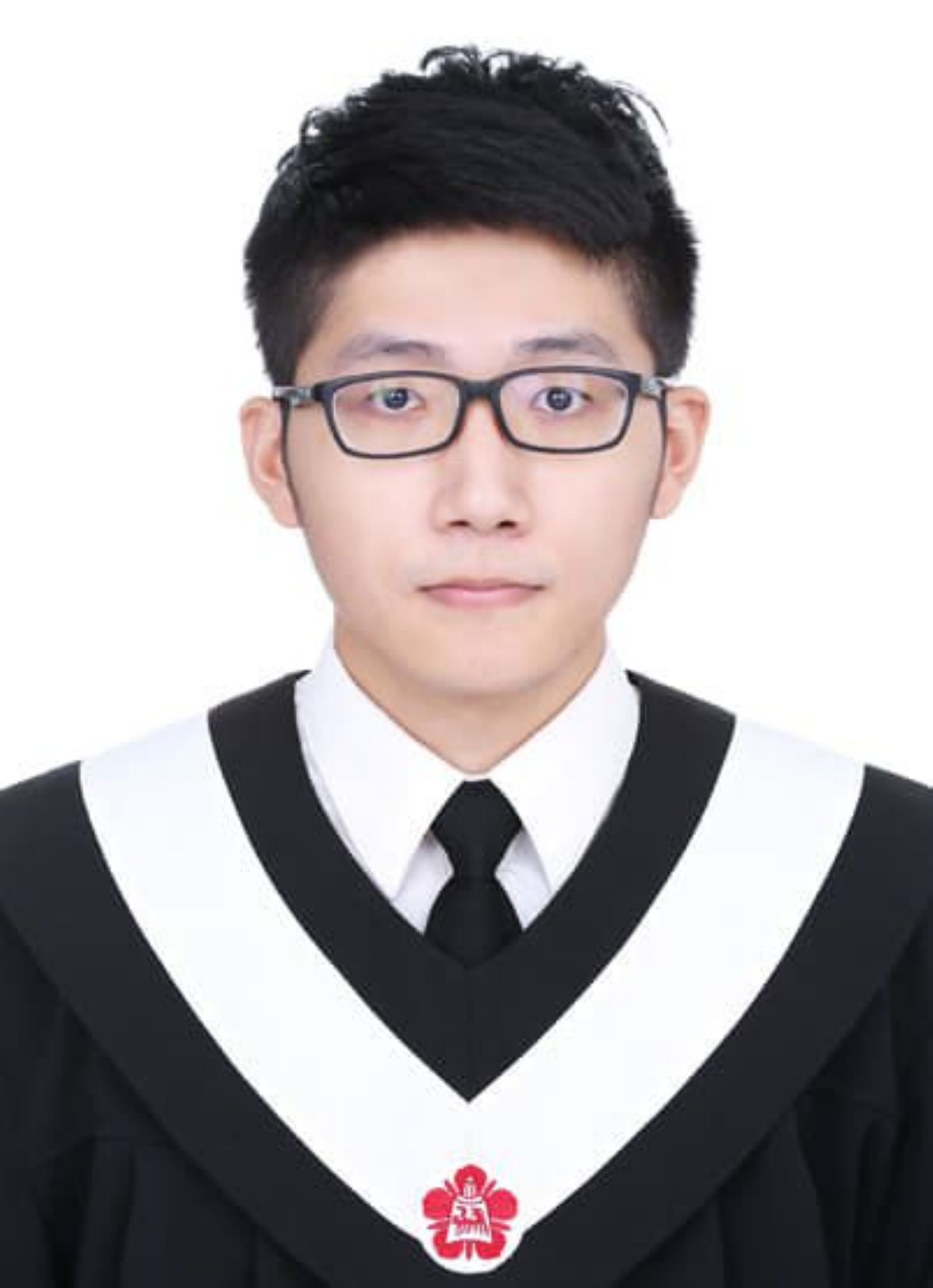}}]{Yu-Cheng Chang}
received his M.S. degree from the Department of Computer Science and Information Engineering, National Taiwan University, Taipei, Taiwan, in 2020. His research interests include machine learning and medical image processing.
\end{IEEEbiography}

\begin{IEEEbiography}
    [{\includegraphics[width=1in,height=1.25in,clip,keepaspectratio]{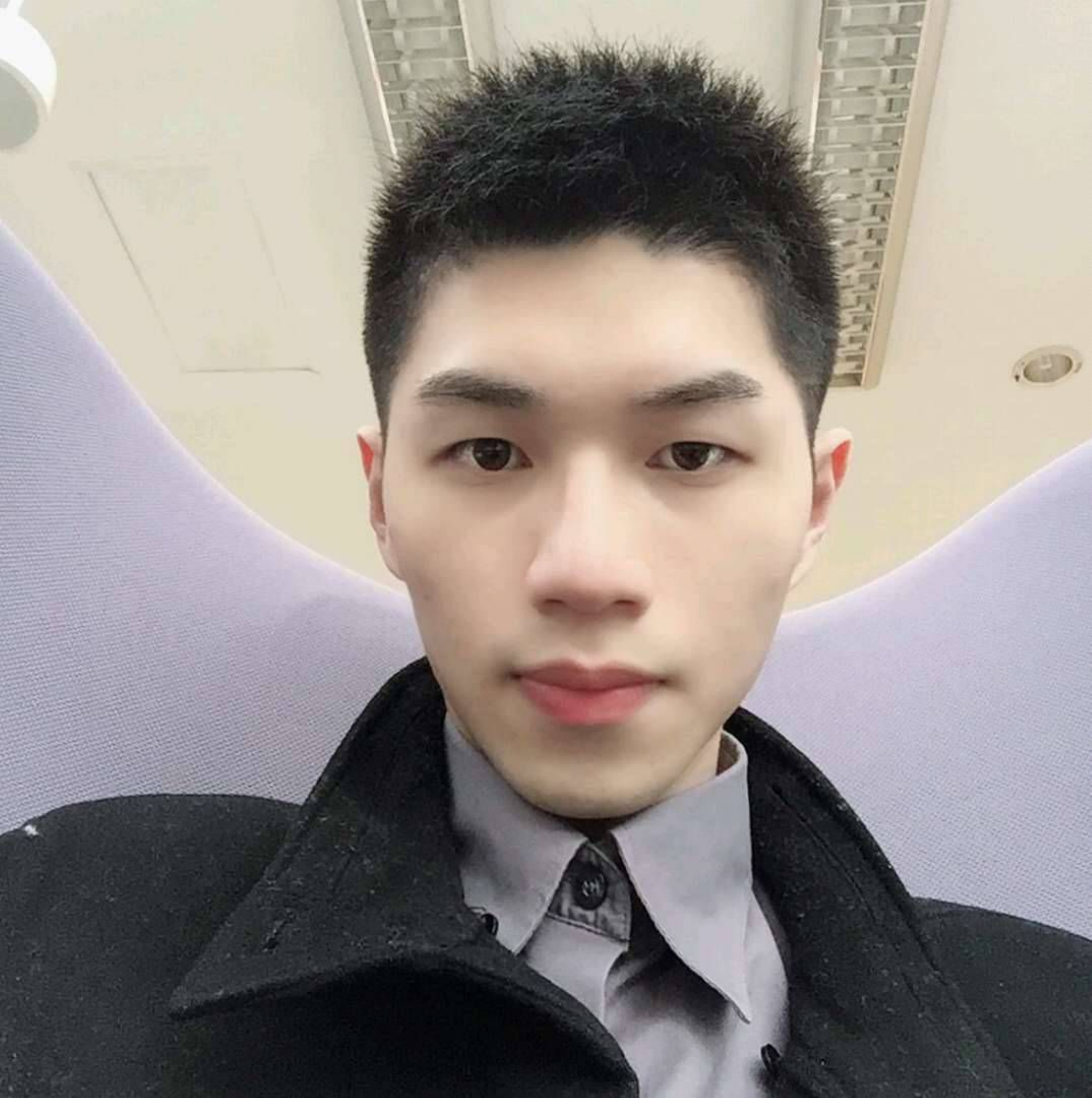}}]{Yu-Hsiang Lin}
 received the M.S. degree in statistics from National Yang Ming Chiao Tung University, Hsinchu, Taiwan, in 2019. He is currently a research assistant at National Tsing Hua University, Taiwan. His research interests include object detection, natural language processing, sentiment analysis, and deep learning.
 \end{IEEEbiography}

\begin{IEEEbiography}
    [{\includegraphics[width=1in,height=1.25in,clip,keepaspectratio]{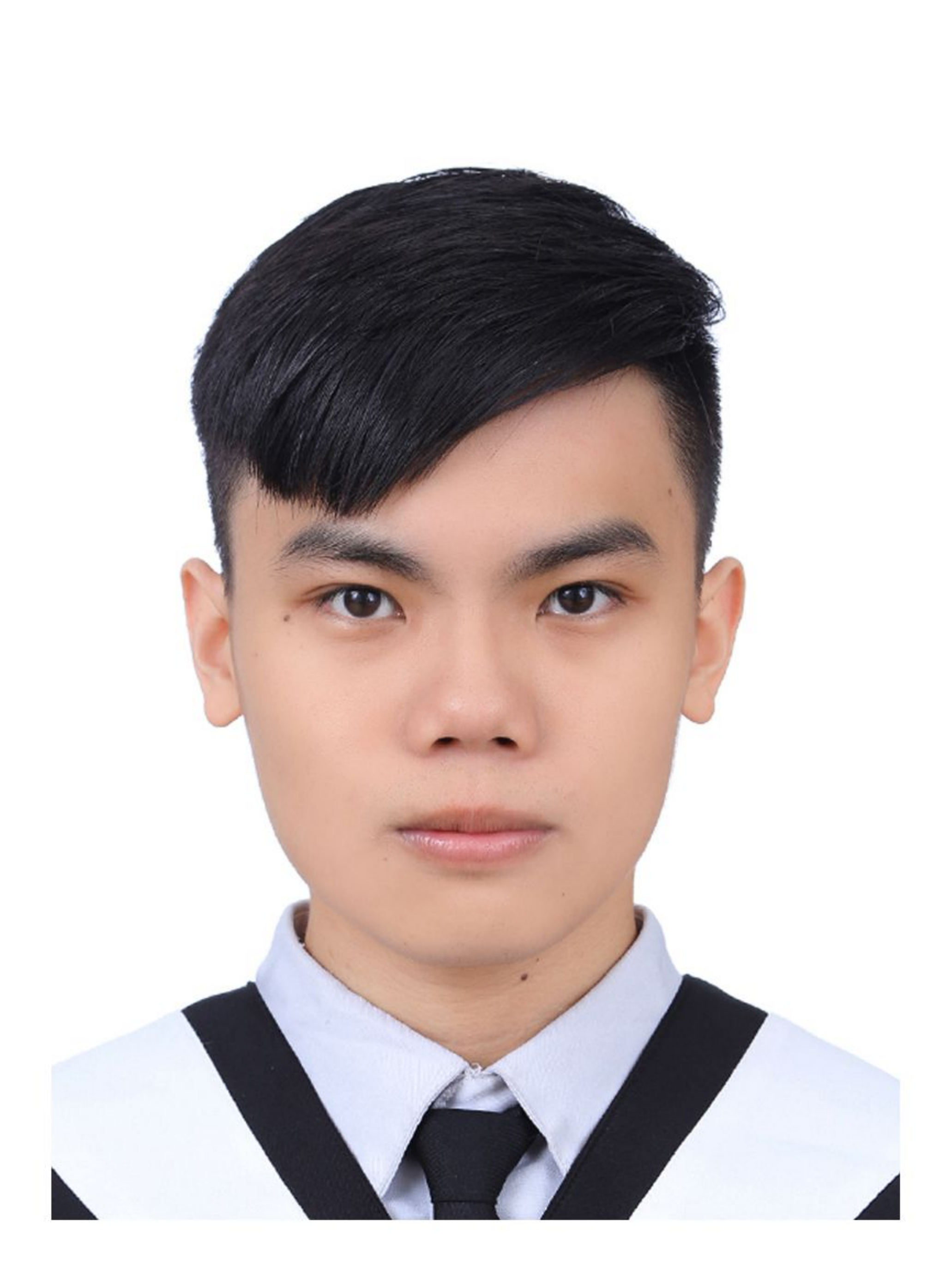}}]{Jia-Fong Yeh}
 received the B.S. degree and the M.S. degree in Computer Science and Information Engineering from National Taiwan Normal University (NTNU), Taiwan, in 2017 and 2019, respectively. He is currently pursuing his Ph.d. degree in Computer Science and Information Engineering at Natioal Taiwan University (NTU), Taiwan. His research interests include machine learning (ML), evolutionary algorithms (EAs), and computer games (CG). Recently, He is devoted to the study on few-shot learning. Jia-Fong is a student member of IEEE.
\end{IEEEbiography}

\begin{IEEEbiography}
[{\includegraphics[width=1in,height=1.25in,clip,keepaspectratio]{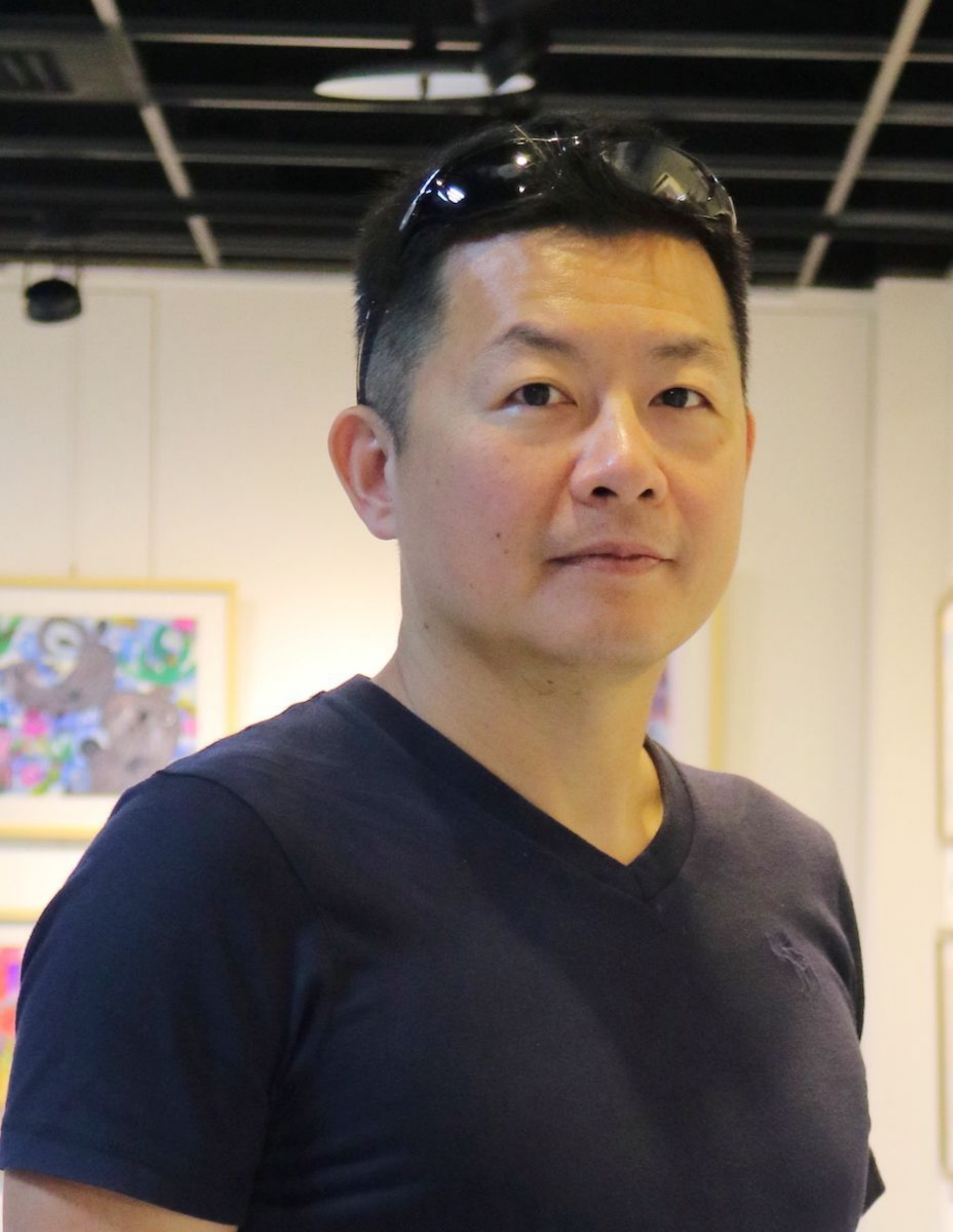}}]{Winston H. Hsu}
(S’03–M’07–SM’12) received the Ph.D. degree in electrical engineering from Columbia University, New York, NY, USA. He is keen to realizing advanced researches towards business deliverables via academia-industry collab- orations and co-founding startups. Since 2007, he has been a Professor with the Graduate Institute of Networking and Multimedia and the Department of Computer Science and Information Engineering, National Taiwan University. His research interests include large-scale image/video retrieval/mining, vi-
sual recognition, and machine intelligence. Dr. Hsu served as the Associate Editor for the IEEE TRANSACTIONS ON MULTIMEDIA and on the Editorial Board for the IEEE MultiMedia Magazine.
\end{IEEEbiography}

\begin{IEEEbiography}
[{\includegraphics[width=1in,height=1.25in,clip,keepaspectratio]{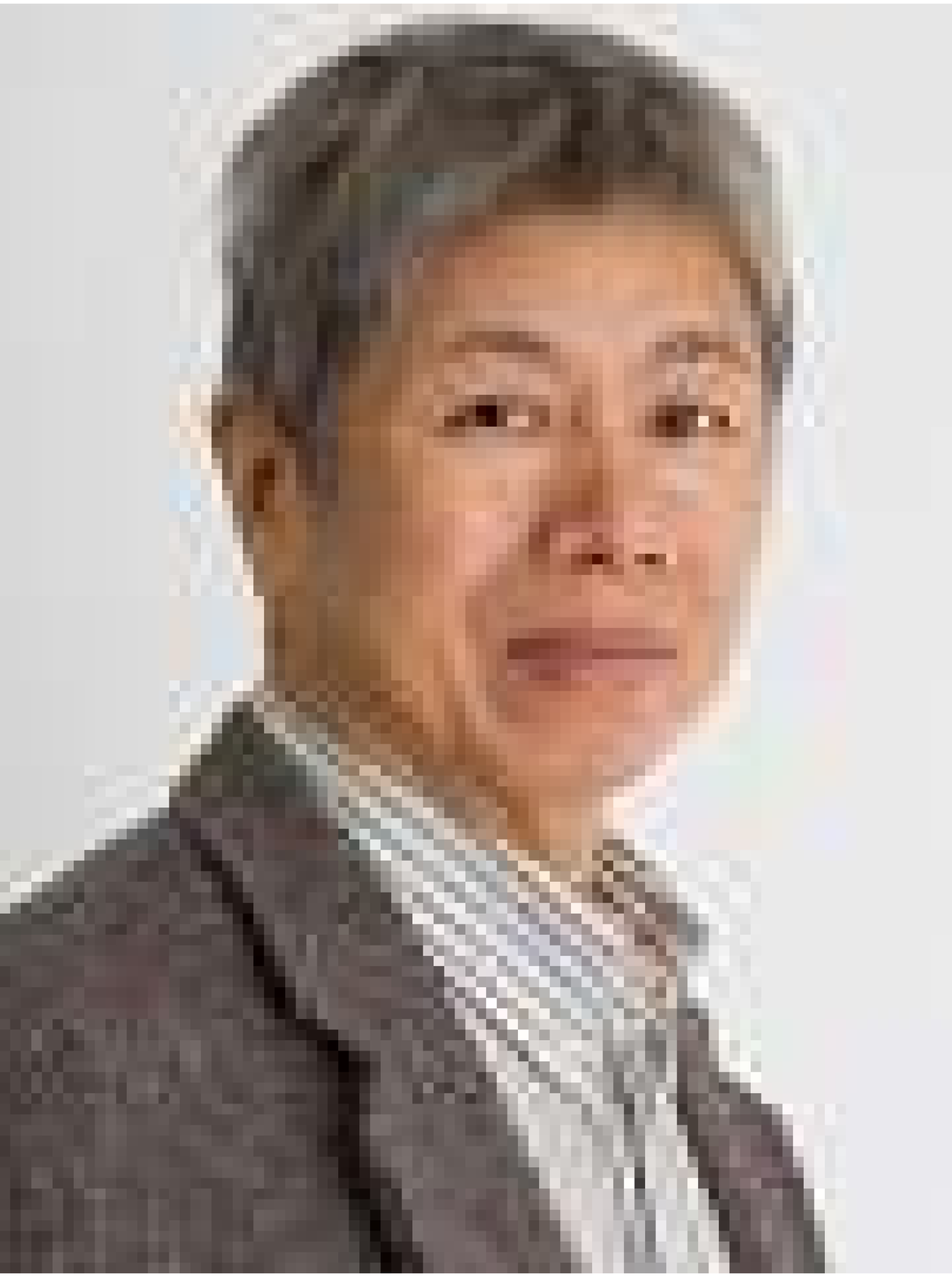}}]{Wen-Chin Chen}
received a BS in mathematics from National Taiwan University and a PhD in compputer science from Brown University in 1976 and 1984.
He has been the Professor of Computer science and Information Engineering department of National Taiwan University since 1987.
His research interests are in the areas of design and analysis of algorithms, multimedia systems and quantumn algorithms.
\end{IEEEbiography}

\vfill


\end{document}